\documentclass[runningheads]{llncs}
\usepackage{graphicx}

\usepackage{tikz}
\usepackage{comment}
\usepackage{amsmath,amssymb} 
\usepackage{color}
\usepackage{graphicx}
\usepackage{subfigure}

\usepackage{booktabs}
\usepackage{algorithm}
\usepackage{algorithmic}
\usepackage{amsmath}
\usepackage{amssymb}
\usepackage{amsmath,amssymb,amsfonts}
\usepackage{algorithm}
\usepackage{booktabs}
\usepackage{algorithmic}
\usepackage{bm}
\usepackage{multirow}
\usepackage{amssymb}
\usepackage{mathtools}
\usepackage{makecell}
\usepackage{enumitem}
\usepackage{bbding}
\usepackage[accsupp]{axessibility}  

\begin{document}
\pagestyle{headings}
\mainmatter
\def\ECCVSubNumber{3827}  

\title{Categories of Response-Based, Feature-Based, and Relation-Based Knowledge Distillation} 

\titlerunning{Knowledge Distillation}
%
\author{Chuanguang Yang\inst{1,2(}\Envelope\inst{)} \and
	Xinqiang Yu\inst{1,2} \and
	Zhulin An\inst{1} \and Yongjun Xu\inst{1}}
\authorrunning{C. Yang et al.}
%
\institute{Institute of Computing Technology, Chinese Academy of Sciences, Beijing, China \and
	University of Chinese Academy of Sciences, Beijing, China \\ 
	\email{\{yangchuanguang, yuxinqiang21s, anzhulin, xyj\}@ict.ac.cn} }

\maketitle

\begin{abstract}
		Deep neural networks have achieved remarkable performance for artificial intelligence tasks. The success behind intelligent systems often relies on large-scale models with high computational complexity and storage costs. The over-parameterized networks are often easy to optimize and can achieve better performance. However, it is challenging to deploy them over resource-limited edge-devices.  Knowledge Distillation (KD) aims to optimize a lightweight network from the perspective of over-parameterized training. The traditional offline KD transfers knowledge from a cumbersome teacher to a small and fast student network. When a sizeable pre-trained teacher network is unavailable, online KD can improve a group of models by collaborative or mutual learning. Without needing extra models, Self-KD boosts the network itself using attached auxiliary architectures. KD mainly involves knowledge extraction and distillation strategies these two aspects. Beyond KD schemes, various KD algorithms are widely used in practical applications, such as multi-teacher KD, cross-modal KD, attention-based KD, data-free KD and adversarial KD. This paper provides a comprehensive KD survey, including knowledge categories, distillation schemes and algorithms, as well as some empirical studies on performance comparison. Finally, we discuss the open challenges of existing KD works and prospect the future directions. 
\end{abstract}
\section{Categories of response-based, feature-based, and relation-based knowledge distillation.} 
\label{Categories}
\begin{figure}[tbp]  
	\centering 
	\includegraphics[width=0.8\linewidth]{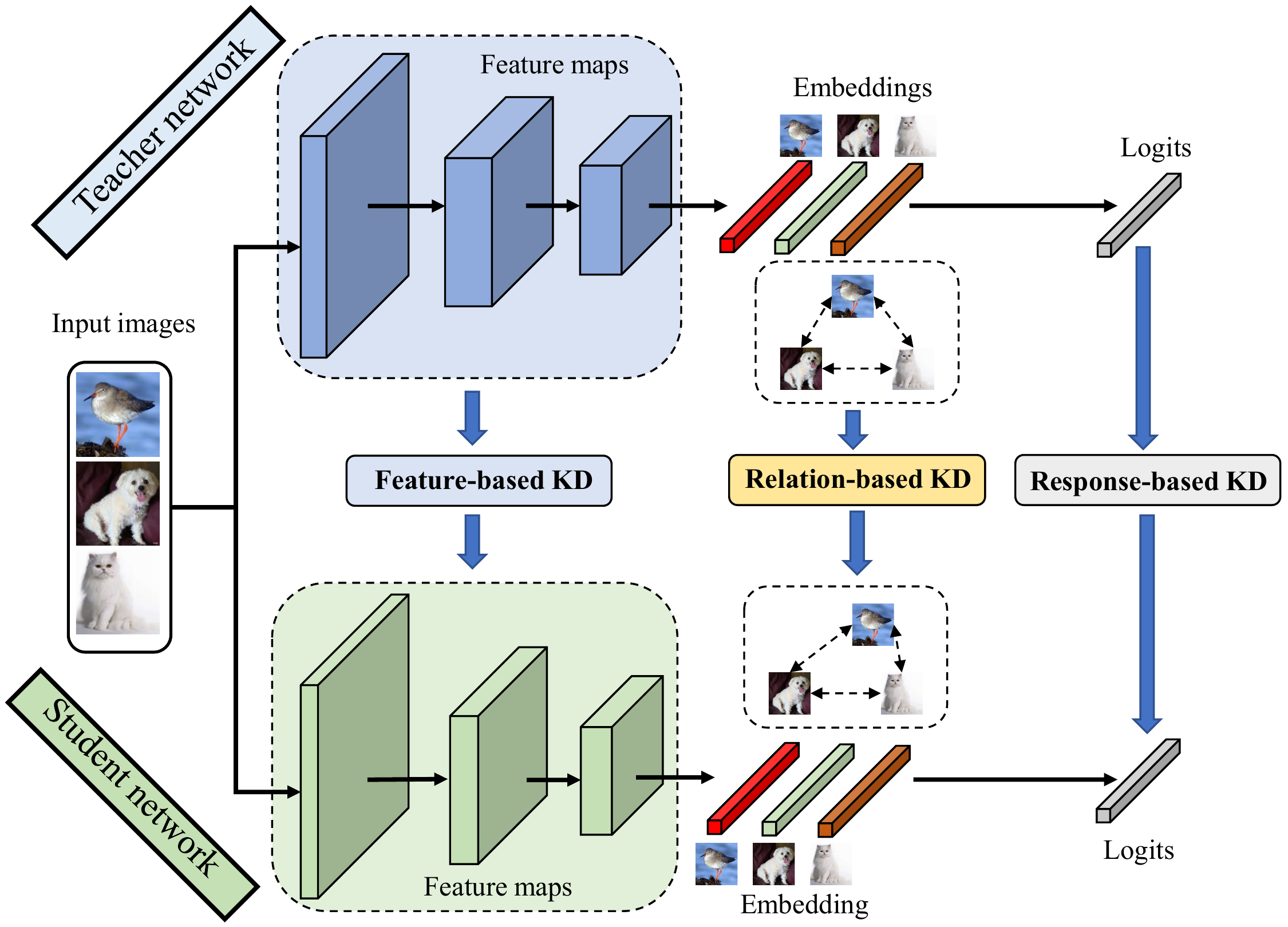}
	\caption{\textbf{The schematic illustration of response-based, feature-based, and relation-based offline KD between teacher and student networks.}} 
	\label{overview_kd}
\end{figure}

The current offline knowledge distillation methods often involve knowledge type and distillation strategies. The former focuses on exploring various information types for student mimicry. The latter aims to help the student to learn teacher effectively. In this section, we investigate response-based, feature-based, and relation-based knowledge, commonly utilized types based on their pre-defined information. Response-based KD guides the teacher's final output to instruct the student's output~\cite{hinton2015distilling,jin2019knowledge,zhou2021rethinking}. This makes intuitive
sense to let the student know how a powerful teacher
produces the predictions. Besides the final output, intermediate features encode the process of knowledge abstract from a neural network. Feature-based KD~\cite{romero2014fitnets,zagoruyko2016paying,yim2017gift,heo2019knowledge,huang2017like} can teach the student to obtain more meaningful semantic information throughout the hidden layers. Response-based and feature-based KD often consider knowledge extraction from a single data sample. Instead, relation-based KD~\cite{yim2017gift,lee2019graph,liu2019knowledge,tung2019similarity,park2019relational,yang2022cross} attempts to excavate cross-sample relationships across the whole dataset. In this section, we survey some representative approaches for each knowledge type and summarize their difference. The overview of schematic illustrations is shown in Fig.~\ref{overview_kd}.

\subsection{Response-based Knowledge Distillation}

Response-based KD focuses on learning knowledge from the last layer as the response. It aims to align the final predictions between the teacher and student. The property of response-based KD is outcome-driven learning that makes it be readily extended to various tasks. The seminal KD dates back to Hinton~\emph{et al.}~\cite{hinton2015distilling}. The core idea is to distill class probability distribution via softened softmax (namely 'soft label'). For the classification task, the soft probability distribution $p$ is formulated as Eq.~\ref{p}:
\begin{equation}
	p(z_{i};T)=\frac{\exp{(z_{i}/T)}}{\sum_{j=1}^{N}\exp{(z_{j}/T)}},
	\label{p}
\end{equation}
where $z_{i}$ is the logit value of the $i$-th classes, $N$ is the number of classes and $T$ is a temperature parameter to adjust the smoothness of class probability distribution. Given the class probability distribution from the teacher as $p(z^{T};T)$ and the student as $p(z^{S};T)$, response-based KD attempts to match $p(z^{S};T)$ with $p(z^{T};T)$ using a distance function denoted as $\mathcal{L}_{dis}$:
\begin{equation}
	\mathcal{L}_{response\_kd}(p(z^{S};T),p(z^{T};T))=\mathcal{L}_{dis}(p(z^{S};T),p(z^{T};T)),
\end{equation}
where $\mathcal{L}_{dis}$ can be formulated as Kullback-Leibler divergence loss~\cite{hinton2015distilling}, Mean Squared Error loss~\cite{kim2021comparing} or pearson correlation coefficient~\cite{huang2022knowledge}. For example, the loss of the conventional KD is formulated as:
\begin{equation}
	\mathcal{L}_{KD}(p(z^{S};T),p(z^{T};T))=\sum_{n=1}^{N}p(z^{T};T)[n]\log\frac{p(z^{T};T)[n]}{p(z^{S};T)[n]},
\end{equation}
where $[n]$ denotes the index of the $n$-th probability value.

\textbf{Interpret response-based KD}. The core idea of response-based KD is straightforward to understand. It guides the student to learn the final results generated from the teacher network. The efficacy of response-based KD can also be connected to label smoothing regularization~\cite{kim2017transferring,ding2019adaptive,muller2019does,yuan2020revisiting,shen2021label}. Yuan~\emph{et al.}~\cite{yuan2020revisiting} resolves the equation of KL-divergence-based KD to label smoothing. Muller~\emph{et al.}~\cite{muller2019does} observed that label smoothing suppresses the effectiveness of KD. They think that label smoothing leads to information loss in the logits on similarities among samples from various classes, which is valuable for response-based KD. However, Shen~\emph{et al.}~\cite{shen2021label} provided an empirical study to illuminate that label smoothing does not suppress KD's effectiveness generally. They found that label smoothing may result in a negative impact under two scenarios: long-tailed class distribution and increased number of classes. Mobahi~\emph{et al.}~\cite{mobahi2020self} showed a theoretical perspective that distillation between two identical network architectures amplifies regularization in Hilbert space. The latest DKD~\cite{zhao2022decoupled} decouples the original KD loss~\cite{hinton2015distilling} into  target class KD (TCKD) and non-target class KD (NCKD). By only introducing two hyper-parameters to balance two terms flexibly, DKD~\cite{zhao2022decoupled} improves the effectiveness of the original KD.

\textbf{Reduce performance gap with auxiliary architecture}. A performance gap may exist in KD due to the capacity gap between teacher and student, leading to a performance degradation issue. To alleviate this problem, TAKD~\cite{mirzadeh2020improved} introduces an intermediate-sized network as a teacher assistant and performs a sequential KD process. Along the vein, HKD~\cite{passalis2020heterogeneous} applies an auxiliary teacher network to transfer layer-wise information flow. DGKD~\cite{son2021densely} proposes a densely guided pattern and performs multi-step KD by all previous teachers. SFTN~\cite{park2021learning} trains a teacher along with student branches at first and then transfers more easy-to-transfer knowledge to the student. TOFD~\cite{zhang2020task} and HSAKD~\cite{yang2021hierarchical} attach several auxiliary branches between teacher and student to facilitate knowledge interaction. However, these methods introduce extra architectures to the training graph and increase training costs.

\textbf{Reduce performance gap with adaptive distillation}. Some works attempted to investigate the sample-aware adaptive distillation to improve performance. WSL~\cite{zhou2021rethinking} proposes sample-wise weighted soft labels from the perspective of the bias-variance tradeoff. ATKD~\cite{guo2021reducing} uses an adaptive temperature based on standard deviation to reduce the sharpness gap between teacher and student. MKD~\cite{liu2022meta} utilizes meta-learning to search a learnable temperature parameter. SCKD~\cite{zhu2021student} examines the mismatch problem from the view of gradient similarity, making the student adaptively learn its beneficial knowledge.  PAD~\cite{zhang2020prime} proposes an adaptive sample weighting mechanism with data uncertainty based on the observation that hard instances may be intractable for KD. Beyond exploring sample-dependent distillation, Song \emph{et al.}~\cite{song2021robust} and Li~\emph{et al.}~\cite{li2020residual} proposed a hybrid forward scheme to make the student learn the teacher's knowledge implicitly via joint training. Inspried by curriculum learning, RCO~\cite{jin2019knowledge} forces the student to mimic training trajectories of the teacher from scratch to convergence. ESKD~\cite{cho2019efficacy} stops the teacher training early to produce more softened logits.

\textbf{Discussion.} We provide an empirical study of various response-based KD approaches in Table~\ref{response}. Different distillation algorithms may be superior to various network architectures. Most methods are applied to convolutional neural networks, while recent MKD~\cite{liu2022meta} further aims to improve vision transformers.  The core idea of response-based KD is outcome-driven learning and readily applicable to existing recognition tasks. For object detection, Chen~\emph{et al.}~\cite{chen2017learning} proposed to guide the student to mimic the teacher's object probabilities and regression bounding boxes. For semantic segmentation, the outcome-driven knowledge is pixel-wise class probability distribution~\cite{liu2019structured}. Analogously, for Bert compression in natural language processing, DistilBERT~\cite{sanh2019distilbert} transfers class predictions of masked tokens. Although response-based KD has been successfully applied to many tasks, the performance gap is still an open issue. When the capacity gap between teacher and student is quite large, the student may not be able to absorb meaningful knowledge. This may lead to adverse supervisory effects. Furthermore, response-based KD ignores the intermediate information encoded in the hidden layers of a neural network, resulting in limited performance improvements.

\begin{table}[tbp]
	\centering
	\caption{Top-1 accuracy(\%) of response-based offline KD methods for ImageNet~\cite{deng2009imagenet} classification. The compared works are sorted according to the published time. The networks are selected as ResNets~\cite{he2016deep}, MobileNets~\cite{sandler2018mobilenetv2}, CaiT~\cite{touvron2021going} and ViT~\cite{dosovitskiy2020image}. All results are referred to the original papers.}
	\resizebox{1.\linewidth}{!}{
		\begin{tabular}{l|c|c|ccc}  
			\toprule
			
			Method&Venue&Algorithm&Teacher(baseline)&Student(baseline)&After KD\\
			\midrule
			KD~\cite{hinton2015distilling}&ArXiv-2015& vanilla KD &ResNet-34(73.3)&ResNet-18(69.8)&70.7\\
			RCO~\cite{jin2019knowledge}&ICCV-2019&curriculum learning &ResNet-50(75.5)&MobileNetV2(64.2)&68.2\\
			TAKD~\cite{mirzadeh2020improved}&AAAI-2020&teacher assistant&ResNet-50(76.1)&ResNet-14(65.2)&67.4\\
			PAD~\cite{zhang2020prime}&ECCV-2020&uncertainty learning &ResNet-34(73.3)&ResNet-18(69.8)&71.7\\
			TOFD~\cite{zhang2020task}&NeurIPS-2020&task-oriented&ResNet-152(78.3)&ResNet-18(69.8)&70.9\\
			DGKD~\cite{son2021densely}&ICCV-2021&densely guidance&ResNet-34(73.3)&ResNet-18(69.8)&71.7\\
			SFTN~\cite{park2021learning}&NeurIPS-2021 &prior training&ResNet-50(77.4)&ResNet-34(73.8)&75.5\\

			HSAKD~\cite{yang2021hierarchical}&IJCAI-2021&self-supervision&ResNet-34(73.3)&ResNet-18(69.8)&72.4\\
			SCKD~\cite{zhu2021student}&ICCV-2021&gradient similarity&ResNet-101(77.4)&ResNet-18(70.3)&71.3\\
			WSL~\cite{zhou2021rethinking}&ICLR-2021&bias-variance tradeoff&ResNet-34(73.3)&ResNet-18(69.8)&72.0\\
			ATKD~\cite{guo2021reducing}&Openreview-2021&sharpness gap&ResNet-34(73.3)&ResNet-18(69.8)&72.8\\
			DKD~\cite{zhao2022decoupled}&CVPR-2022& balancing losses &ResNet-34(73.3)&ResNet-18(69.8)&71.7\\
			MKD~\cite{liu2022meta}&Arxiv-2022
			&meta-learning& CaiT-S24(82.4)&ViT-T(72.2)&76.4\\
			\bottomrule
	\end{tabular}}
	
	\label{response} 
\end{table}

\subsection{Feature-based Knowledge Distillation} 
As we discussed above, response-based KD neglects intermediate-level supervision for complete guidance. To address this defect, feature-based KD focuses on exploring intermediate feature information to provide comprehensive supervisory, such as feature maps and their refined information. The common feature-based distillation loss can be formulated as Eq.~\ref{feature}:
\begin{equation}
\mathcal{L}_{feature\_kd}(F^{S},F^{T})=\mathcal{L}_{dis}{(\phi^{S}(F^{S}),\phi^{T}(F^{T}))},
\label{feature}
\end{equation}
where $F^{S}$ and $F^{T}$ represent intermediate feature maps from student and teacher. $\phi^{S}$ and $\phi^{T}$ are meaningful transformation functions to produce refined information, such as attention mechanism~\cite{komodakis2017paying}, activation boundary~\cite{heo2019knowledge}, neuron selectivity~\cite{huang2017like} and probability distribution~\cite{passalis2018learning}, etc. $\mathcal{L}_{dis}$ is a distance function that measures the similarity of matched feature information, for example, Mean Squared Error loss~\cite{romero2014fitnets,komodakis2017paying,heo2019knowledge} and Kullback-Leibler divergence loss~\cite{passalis2018learning}. For example, the seminal FitNet~\cite{romero2014fitnets} is formulated as:
\begin{equation}
\mathcal{L}_{FitNet}(F^{S},F^{T})=\frac{1}{H\times W\times C}\sum_{h=1}^{H}\sum_{w=1}^{W}\sum_{c=1}^{C}(F^{S}[h,w,c]-F^{T}[h,w,c])^{2}.
\end{equation}
Here, we assume $F^{S}\in \mathbb{R}^{H\times W\times C}$ and $F^{T}\in \mathbb{R}^{H\times W\times C}$, where $H$, $W$, $C$ denote the feature map's height, weight and channel number, respectively.

\textbf{Knowledge exploration: transform intermediate feature maps to meaningful knowledge.} The seminal FitNet~\cite{romero2014fitnets} is the first feature-based KD method. Its core idea is to align the intermediate feature maps generated from the hidden layers in a layer-by-layer manner between teacher and student. This simple and intuitive work may not use high-level knowledge. Subsequent approaches attempted to explore more meaningful information encoded in the raw feature maps that is more suitable for feature-based KD.

 AT~\cite{komodakis2017paying} transforms the feature map to a spatial attention map as valuable information. NST~\cite{huang2017like} extracts activation heatmaps as neuron selectivity for transfer. Srinivas \emph{et al.}~\cite{srinivas2018knowledge} applied Jacobian matching between feature maps. PKT~\cite{passalis2018learning} formulates the feature map as a probability distribution and mimicked by KL-divergence. FSP~\cite{yim2017gift} introduces Gramian matrix~\cite{gatys2015neural} to measure the flow of solution procedure across feature maps from various layers. Seung \emph{et al.}~\cite{lee2018self} used singular value decomposition to resolve the feature knowledge. FT~\cite{kim2018paraphrasing} introduces an auto-encoder to parse the teacher's feature map as "factors" in an unsupervised manner and a translator to transform "factors" to easily understandable knowledge. AB~\cite{heo2019knowledge} considers activation boundaries in the hidden feature space and forces the student to learn consistent boundaries with the teacher. Overhaul~\cite{heo2019comprehensive} rethinks the distillation feature position with a newly designed margin ReLU and a partial L2 distance function to filter redundant information.
 
  More recently, TOFD~\cite{zhang2020task} and HSAKD~\cite{yang2021hierarchical} attach auxiliary classifiers to intermediate feature maps supervised by extra tasks to produce informative probability distributions. The former leverages the original supervised task, while the latter introduces a meaningful self-supervised augmented task. MGD~\cite{yue2020matching} performs fine-grained feature distillation with an adaptive channel assignment algorithm. ICKD~\cite{liu2021exploring} excavates inter-channel correlation from feature maps containing the feature space's diversity and homology. Beyond the channel dimension, TTKD~\cite{lin2022knowledge} conducts spatial-level feature matching using self-attention mechanism~\cite{vaswani2017attention}. In summary, previous methods often resort to extracting richer feature information for KD, leading to better performance than vanilla FitNet~\cite{romero2014fitnets}.

 \textbf{Knowledge transfer: good mimicry algorithm to let the student learn better.} Beyond knowledge exploration, another valuable problem is how to transfer knowledge effectively. Most feature-based KD approaches use the simple Mean Squared Error loss for knowledge alignment. Besides this vein, VID~\cite{ahn2019variational} refers to the information-theoretic framework and considers KD as maximizing the mutual information between teacher and student. Wang~\emph{et al.}~\cite{wang2018adversarial} regarded the student as a generator and applied an extra discriminator to distinguish features produced from student or teacher. This adversarial process guides the student to learn the similar feature distribution to the teacher. Xu~\emph{et al.}~\cite{xu2020feature} proposed to normalize feature embeddings in penultimate layer to suppress the negative impact of noise. Beyond examining mimicry metric loss, using a shared classifier between teacher and student can also help the student to align the teacher's features implicitly~\cite{yang2021knowledge,chen2022knowledge}. 
 
 \begin{table}[tbp]
 	\centering
 	\caption{Top-1 accuracy(\%) of various feature-based offline KD methods for CIFAR-100~\cite{krizhevsky2009learning} classification. The compared works are sorted according to the published time. The networks are selected as ResNets~\cite{he2016deep} and WRNs~\cite{zagoruyko2016wide}. All results are referred to the original papers.}
 	\resizebox{1.\linewidth}{!}{
 		\begin{tabular}{l|c|c|ccc}  
 			\toprule
 			
 			Method&Venue&Knowledge&Teacher(baseline)&Student(baseline)&After KD\\
 			\midrule
 			FitNet~\cite{romero2014fitnets}&ICLR-2015&feature maps&ResNet-56(72.34)&ResNet-20(69.06)&69.21\\
 			AT~\cite{komodakis2017paying}&ICLR-2017&attention maps&ResNet-56(72.34)&ResNet-20(69.06)&70.55\\
 			FSP~\cite{yim2017gift}&CVPR-2017&solution flow&ResNet-56(72.34)&ResNet-20(69.06)&69.95\\
 			NST~\cite{huang2017like}&arXiv-2017&neuron selectivity&ResNet-56(72.34)&ResNet-20(69.06)&69.60\\
 			Jacobian~\cite{srinivas2018knowledge}&ICML-2018&gradient&WRN-28-4(78.91)&WRN-16-4(77.28)&77.82\\
 			FT~\cite{kim2018paraphrasing}&NeurIPS-2018&paraphrased factor&ResNet-56(72.34)&ResNet-20(69.06)&69.84\\
 			PKT~\cite{passalis2018learning}&ECCV-2018&probability distribution&ResNet-56(72.34)&ResNet-20(69.06)&70.34\\
 			AB~\cite{heo2019knowledge}&AAAI-2019&activation boundaries&ResNet-56(72.34)&ResNet-20(69.06)&69.47\\
 			VID~\cite{ahn2019variational}&CVPR-2019&mutual information&ResNet-56(72.34)&ResNet-20(69.06)&70.38\\
 			Overhaul~\cite{heo2019comprehensive}&ICCV-2019&feature position&WRN-28-4(78.91)&WRN-16-4(77.28)&79.11\\
 			FKD~\cite{xu2020feature}&ECCV-2020&normalization&ResNet-56(81.73)&ResNet-20(78.30)&81.19\\
 			DFA~\cite{guan2020differentiable}&ECCV-2020&differentiable search &WRN-28-4(79.17)&WRN-16-4(77.24)&79.74\\
 			MGD~\cite{yue2020matching}&ECCV-2020&channel assignment&WRN-28-4(78.91)&WRN-16-4(77.28)&78.88\\
 			TOFD~\cite{zhang2020task}&NeurIPS-2020&task-oriented&ResNet-56(73.44)&ResNet-20(69.6)&72.02\\
 			HSAKD~\cite{yang2021hierarchical}&IJCAI-2021&self-supervision&ResNet-56(73.44)&ResNet-20(69.6)&72.60\\
 			ICKD~\cite{liu2021exploring}&ICCV-2021&inter-channel correction &ResNet-56(72.34)&ResNet-20(69.06)&71.76\\
 			SRRL~\cite{yang2021knowledge}&ICLR-2021&softmax regression&WRN-40-2(76.31)&WRN-40-1(71.92)&74.64\\
 			SimKD~\cite{chen2022knowledge}&CVPR-2022&reused classifier&WRN-40-2(76.31)&WRN-40-1(71.92)&75.56\\
 			TTKD~\cite{lin2022knowledge}&CVPR-2022&spatial attention&ResNet-56(72.34)&ResNet-20(69.06)&71.59\\

 			\bottomrule
 	\end{tabular}}
 	
 	\label{feature_kd} 
 \end{table}

	\textbf{Distillation for vision transformer.} Vision transformer (ViT)~\cite{dosovitskiy2020image} has shown predominant performance for image recognition. However, ViT-based networks need high demand of computational costs. KD provides an excellent solution to train a small ViT with desirable performance. Over the relation level, Manifold Distillation~\cite{jia2021efficient} explores patch-wise relationships as the knowledge type for ViT KD. Over the feature level, AttnDistill~\cite{wang2022attention} transfers attention maps from teacher to student. ViTKD~\cite{yang2022vitkd} provides practical guidelines for ViT feature distillation. Besides feature mimicry between homogeneous ViTs, some works~\cite{zhang2022bootstrapping,chen2022dearkd} also attempt to distill inductive biases from CNN to ViT. Some promising knowledge types are still worth further mining, such as intermediate features, attentive relationships and distillation positions.

  \textbf{Discussion.} We provide an empirical study of various feature-based KD approaches in Table~\ref{feature_kd}. Generally, feature-based KD is a comprehensive supplement to response-based KD that provides intermediate features encapsulating the learning process. 
  However, simply aligning the same-staged feature information between teacher and student may result in negative supervisory, especially when the capacity gap or architectural difference is large. A more valuable direction may lie in the student-friendly feature-based KD that provides semantic-consistent supervision.

\subsection{Relation-based Knowledge Distillation}
Response-based and feature-based KD often consider distilling knowledge from individual samples. Instead, relation-based KD explores \textbf{cross-sample} or \textbf{cross-layer} relationships as meaningful knowledge.

\subsubsection{Relation-based Cross-Sample Knowledge Distillation}
~\\
A general relation-based cross-sample distillation loss is formulated as Eq.~\ref{relation}:
\begin{equation}
\mathcal{L}_{relation\_kd}(F^{S},F^{T})=\sum_{i,j}\mathcal{L}_{dis}{(\psi^{S}(v^{S}_{i},v^{S}_{j}),\psi^{T}(v^{T}_{i},v^{T}_{j}))},
\label{relation}
\end{equation} 
where  $F^{S}$ and $F^{T}$ denote the feature sets of teacher and student, respectively. $v_{i}$ and $v_{j}$ are feature embeddings of the $i$-th and $j$-th samples, and $(v^{S}_{i},v^{S}_{j})\in F^{S}$, $(v^{T}_{i},v^{T}_{j})\in F^{T}$. $\psi^{S}$ and $\psi^{T}$ are similarity metric functions of $(v^{S}_{i},v^{S}_{j})$ and $(v^{T}_{i},v^{T}_{j})$. $\mathcal{L}_{dis}$ is a distance function that measures the similarity of an instance graph, for example, Mean Squared Error loss~\cite{park2019relational,peng2019correlation} and Kullback-Leibler divergence loss~\cite{xu2020knowledge,yang2022cross}. For example, the loss of the representative RKD~\cite{park2019relational} is formulated as:
\begin{align}
	&\mathcal{L}_{RKD}(F^{S},F^{T})=\sum_{i,j,k}(cos\angle v^{S}_{i}v^{S}_{j}v^{S}_{k}-cos\angle v^{T}_{i}v^{T}_{j}v^{T}_{k})^2, \\
	&cos\angle v_{i}v_{j}v_{k}=\left \langle \textbf{e}^{ij},\textbf{e}^{kj} \right \rangle=\left \langle \frac{v_{i}-v_{j}}{\left \| v_{i}-v_{j} \right \|_{2} },\frac{v_{k}-v_{j}}{\left \| v_{k}-v_{j} \right \|_{2} } \right \rangle.
\end{align}

\textbf{Constructing relational graph with various edge weights.} The knowledge of relation-based KD can be seen as an instance graph, where the nodes denote the feature embeddings of samples. Most relation-based KD examines various similarity metric functions to compute edge weights. DarkRank~\cite{chen2018darkrank} is the first approach to examine cross-sample similarities based on Euclidean distance of embeddings for deep metric learning. MHGD~\cite{lee2019graph} processes graph-based representations using a multi-head attention network. RKD~\cite{park2019relational} uses distance-wise and angle-wise similarities of mutual relations as structured knowledge. CCKD~\cite{peng2019correlation} captures the correlation between instances using kernel-based Gaussian RBF. SP~\cite{tung2019similarity} constructs pairwise similarity matrices given the mini-batch. IRG~\cite{liu2019knowledge} models an instance relationship graph with vertex and edge transformation. REFILLED~\cite{ye2020distilling} forces the teacher to reweight the hard triplets forwarded by the student for relationship matching. All methods focus on modeling relation graphs over sample-level feature embeddings but differ in various edge-weight generation strategies.

\textbf{Constructing relational graph with meaningful transformation.} Directly modelling edge-weights using simple metric functions may not capture correlations or higher-order dependencies meaningfully. CRD~\cite{tian2019contrastive} introduces supervised contrastive learning among samples based on an InfoNCE~\cite{oord2018representation}-inspired loss to align the teacher's representations. Over CRD, CRCD~\cite{zhu2021complementary} proposes complementary relation contrastive distillation according to the feature and its gradient. To extract richer knowledge upon the original supervised learning, SSKD~\cite{xu2020knowledge} follows the SimCLR~\cite{chen2020simple} framework and utilizes self-supervised contrastive distillation from image rotations. To take advantage of category-level information from labels, CSKD~\cite{chen2020improving} builds intra-category and inter-category structured relations. Previous methods often focus on instance-level features and their relationships but ignore local features and details. Therefore, LKD~\cite{li2020local} utilizes a class-aware attention module to capture important regions and then models the local relational matrices using the localized patches. GLD~\cite{kim2021distilling} constructs a relational graph with local features extracted by a local spatial pooling layer. 

\subsubsection{Relation-based Cross-Layer Knowledge Distillation}
~\\
Beyond build relationships over data samples, the cross-layer interactive information encoding inside the models is also a valuable knwoledge form. A general relation-based cross-layer distillation loss is formulated as Eq.~\ref{relationx}:

\begin{equation}
\mathcal{L}_{relation\_kd}(f^{S},f^{T})=\mathcal{L}_{dis}{(g^{S}(f^{S}_{i},f^{S}_{j}),g^{T}(f^{T}_{i},f^{T}_{j}))},
\label{relationx}
\end{equation} 
where  $f^{S}$ and $f^{T}$ denote the feature sets extracted from different layers of teacher and student, respectively. $f_{i}$ and $f_{j}$ are feature embeddings from the $i$-th and $j$-th layer, and $(f^{S}_{i},f^{S}_{j})\in f^{S}$, $(f^{T}_{i},f^{T}_{j})\in f^{T}$. $g^{S}$ and $g^{T}$ are layer aggregation functions of $(f^{S}_{i},f^{S}_{j})$ and $(f^{T}_{i},f^{T}_{j})$. $\mathcal{L}_{dis}$ is a distance function that measures the similarity of the cross-layer aggregated feature maps, for example, Mean Squared Error loss~\cite{yim2017gift,chen2021cross,ji2021show,passban2021alp}. For example, the loss of FSP~\cite{yim2017gift} is formulated as:
\begin{align}
	\mathcal{L}_{FSP}(f^{S},f^{T})=\left \| G^{S}(f^{S}_{i},f^{S}_{j})-G^{T}(f^{T}_{i},f^{T}_{j}) \right \|_{2}^{2}, \\
	G(f_{i},f_{j})[a,b]=\sum_{h=1}^{H}\sum_{w=1}^{W}\frac{f_{i}[h,w,a]\times f_{j}[h,w,b]}{H\times W},G(f_{i},f_{j})\in \mathbb{R}^{X\times Y}.
\end{align}
Here, we assume $f_{i}\in \mathbb{R}^{H\times W\times X}$ and $f_{j}\in \mathbb{R}^{H\times W\times Y}$, where $H$ and $W$ denote the height and width, and $X$ and $Y$ represent the number of channels.

\begin{table}[tbp]
	\centering
	\caption{Top-1 accuracy(\%) of various relation-based offline KD methods for CIFAR-100~\cite{krizhevsky2009learning} classification. The compared works are sorted according to the published time. The networks are selected as ResNets~\cite{he2016deep} and WRNs~\cite{zagoruyko2016wide}. All results are referred to the original papers.}
	\resizebox{1.\linewidth}{!}{
		\begin{tabular}{l|c|c|ccc}  
			\toprule
			
			Method&Venue&Knowledge&Teacher(baseline)&Student(baseline)&After KD\\
			\midrule
			MHGD~\cite{lee2019graph}&BMVC-2019&graph attention&ResNet-56(72.34)&ResNet-20(69.06)&69.42\\
			RKD~\cite{park2019relational}&CVPR-2019&relational graph &ResNet-56(72.34)&ResNet-20(69.06)&69.61\\
			IRG~\cite{liu2019knowledge}&CVPR-2019 &instance graph&ResNet-110(72.53)&ResNet-20(68.75)&69.87\\
			CCKD~\cite{peng2019correlation}&ICCV-2019&correlation congruence&ResNet-56(72.34)&ResNet-20(69.06)&69.63\\
			SP~\cite{tung2019similarity}&ICCV-2019&similarity-preserving&ResNet-56(72.34)&ResNet-20(69.06)&69.67\\
			REFILLED\cite{ye2020distilling}&CVPR-2020&relationship matching&WRN-40-2(74.44)&WRN-16-2(70.15)&74.01 \\
			CRD~\cite{tian2019contrastive}&ICLR-2020 &contrastive learning&ResNet-56(72.34)&ResNet-20(69.06)&71.16\\
			SSKD~\cite{xu2020knowledge}&ECCV-2020&self-supervision&ResNet-56(73.44)&ResNet-20(69.63)&71.49\\
			LKD~\cite{xu2020knowledge}&ECCV-2020 &local correlation&ResNet-110(75.76)&ResNet-20(69.47)&72.63\\ 
			SemCKD~\cite{chen2021cross}&AAAI-2021&attention matching&ResNet-32x4(79.42)&ResNet-8x4(73.09)&76.23\\
			AFD~\cite{ji2021show}&AAAI-2021&attention matching&ResNet-56(72.54)&ResNet-20(69.40)&71.53\\
			ReviewKD~\cite{chen2021distilling}&CVPR-2021& feature aggregation&ResNet-56(72.54)&ResNet-20(69.40)&71.89\\
			LONDON~\cite{shang2021lipschitz}&CVPR-2021&Lipschitz continuity &WRN-28-4(78.91)&WRN-16-4(77.28)&79.67\\
			CRCD~\cite{zhu2021complementary}&CVPR-2021 &complementary relation&ResNet-56(72.34)&ResNet-20(69.06)&73.21\\
			GLD~\cite{kim2021distilling}&ICCV-2021&local relationships&ResNet-110(72.53)&ResNet-20(68.75)&71.37 \\
			
			\bottomrule
	\end{tabular}}
	
	\label{relation_kd} 
\end{table}

FSP~\cite{yim2017gift} is the seminal method to capture relationships among cross-layer feature maps for KD. It introduces Gramian matrix~\cite{gatys2015neural} to represent the flow of solution procedure as knowledge. Passalis~\emph{et al.}~\cite{passalis2020heterogeneous} pointed out that the same-staged intermediate layers between teacher and student networks with different capacities may show semantic abstraction gaps. Previous methods~\cite{romero2014fitnets,komodakis2017paying,heo2019knowledge,heo2019comprehensive} often rely on a hand-crafted layer assignment strategy in a one-to-one manner. However, the naive alignment could result in the semantic mismatch problem between the pair-wise teacher-student layers. Many subsequent works consider modeling meaningful knowledge across multiple feature layers. Jang~\emph{et al.}~\cite{jang2019learning} introduced meta-networks for weighted layer-level feature matching. Motivated by self-attention mechanism~\cite{vaswani2017attention}, some works~\cite{chen2021cross,ji2021show,passban2021alp} utilize the attention-based weights for adaptive layer assignment. Apart from examining the layer matching issue, some works~\cite{chen2021distilling,shang2021lipschitz} attempted to aggregate all-staged feature maps to construct informative features as supervisory signals. ReviewKD~\cite{chen2021distilling} leverages multi-level features from the teacher to guide each layer of the student according to various feature fusion modules. LONDON~\cite{shang2021lipschitz} summarizes multi-level feature maps to model Lipschitz continuity. Besides manual strategies, DFA~\cite{guan2020differentiable} applies a  search method to find appropriate feature aggregations automatically.

\textbf{Discussion.} We provide an empirical study of various relation-based KD approaches in Table~\ref{relation_kd}. Independent from feature-based and response-based KD, relation-based methods aim to capture high-order relationships among various samples or different layers. The relational graph captures structured dependencies across the whole dataset. The cross-layer feature relationships encode the information of semantic process. How to model a better relationships using more meaningful node transformation and metric functions or aggregating appropriate layer information are still core problems to be further researched.

\begin{figure}[tbp] 
	\centering 
	\subfigure[Offline KD]{  
		\label{offline_kd} 
		\includegraphics[width=0.29\textwidth]{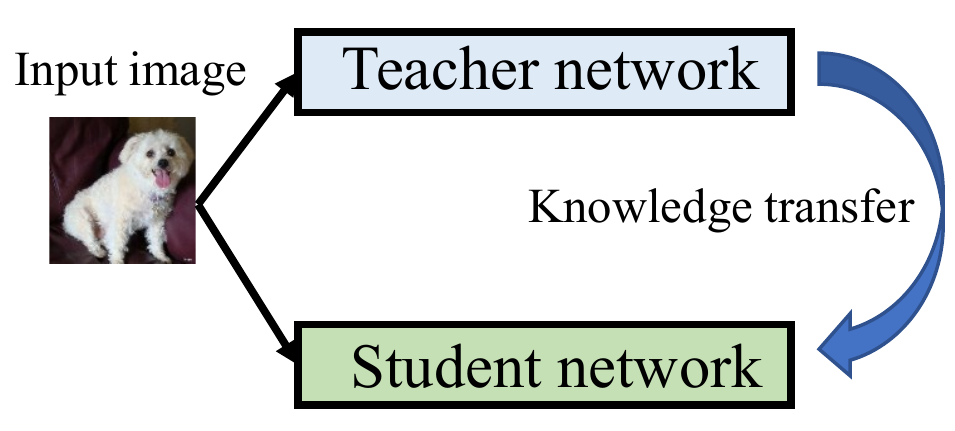}} 
	\subfigure[Online KD]{ 
		\label{online_kd1} 
		\includegraphics[width=0.29\textwidth]{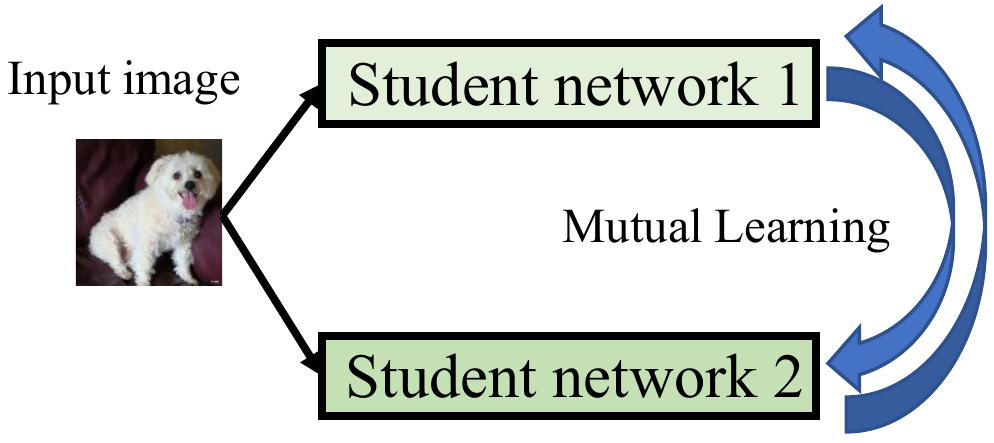}} 
	\subfigure[Self-KD]{ 
		\label{self-kd} 
		\includegraphics[width=0.35\textwidth]{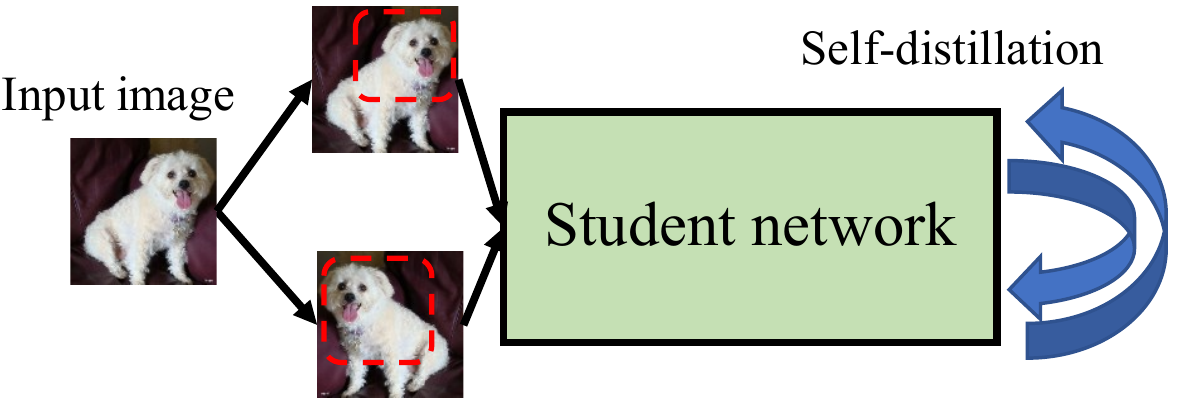}} 
	\caption{\textbf{The schematic illustrations of three KD schemes.} (a) Offline KD performs unidirectional knowledge transfer from a teacher network to a student network. (b) Online KD conducts mutual learning between two peer student networks. (c) Self-KD creates two input views and regularizes similar outputs over a single student network.} 
	\label{kd_scheme} 
\end{figure}

\section{Distillation Schemes}
We discuss distillation schemes of student learning, including offline KD, online KD and Self-KD. The schematic illustrations of three KD schemes are shown in Fig.~\ref{kd_scheme}.

\begin{table}[tbp]
	\centering
	\caption{Comprehensive comparison of representative offline KD methods toward accuracy and distillation time on CIFAR-100 classification when a pre-trained teacher network is available. The distillation time is evaluated on a single NVIDIA Tesla V100 GPU, and is measured from the actual time per epoch. FLOPs denote the number of floating-point operations, measuring the computational complexity of networks. }
	\resizebox{1.\linewidth}{!}{
		\begin{tabular}{l|c|ccc|ccc|c|c}  
			\toprule
			
			\multirow{2}{*}{Method}&\multirow{2}{*}{Venue}&\multicolumn{3}{c|}{Teacher:ResNet-56}&\multicolumn{3}{c|}{Student:ResNet-20}&\multirow{2}{*}{After KD}&\multirow{2}{*}{Time(s)}\\
			&&params&FLOPs&baseline&params&FLOPs&baseline&&\\
			\midrule
			KD~\cite{hinton2015distilling}&ArXiv-2015 &\multirow{13}{*}{0.86M}&\multirow{13}{*}{125.8M}&\multirow{13}{*}{72.34}&\multirow{13}{*}{0.28M}&\multirow{13}{*}{40.8M}&\multirow{13}{*}{69.06}&70.66&20.1\\
			
			FitNet~\cite{romero2014fitnets}&ICLR-2015&&&&&&&69.21&23.5\\
			AT~\cite{komodakis2017paying}&ICLR-2017&&&&&&&70.55&23.2\\

			PKT~\cite{passalis2018learning}&ECCV-2018&&&&&&&70.34&21.4\\
			VID~\cite{ahn2019variational}&CVPR-2019&&&&&&&70.38&27.0\\
			
			RKD~\cite{park2019relational}&CVPR-2019 &&&&&&&69.61&22.7\\
			CCKD~\cite{peng2019correlation}&ICCV-2019&&&&&&&69.63&23.4\\
			CRD~\cite{tian2019contrastive}&ICLR-2020 &&&&&&&71.16&32.5\\
			SSKD~\cite{xu2020knowledge}&ECCV-2020&&&&&&&71.49&33.0\\
			SemCKD~\cite{chen2021cross}&AAAI-2021&&&&&&&71.54&54.0\\
			HSAKD~\cite{yang2021hierarchical}&IJCAI-2021&&&&&&&72.25&75.2\\
			DKD~\cite{zhao2022decoupled}&CVPR-2022&&&&&&&71.97&27.0\\
			
			\bottomrule
	\end{tabular}}
	
	\label{kd_time} 
\end{table}

\subsection{Offline Knowledge Distillation}
The offline KD is so-called teacher-student-based learning~\cite{hinton2015distilling,romero2014fitnets}, which previous works have broadly examined. The core idea of offline KD is to transfer knowledge from a large pre-trained teacher network with high performance to a small and fast student network. In practice, offline KD often conducts a two-stage training pipeline: (1) the teacher network is pre-trained over the task to achieve excellent performance; and (2) the student is guided to mimic the teacher's information during the training phase. When offline KD uses publicly available pre-trained models at hand for training a student network, offline KD can also be regarded as a one-stage pipeline. Because the teacher network is pre-trained and frozen, we call the teacher-student-based learning as offline KD, which is discussed in Section~\ref{Categories} in detail, according to different transferred knowledge types. 

\textbf{Trade-off between performance and distillation time}. We comprehensively compare representative offline KD methods toward accuracy and distillation time. We can observe that various KD approaches have different properties. The conventional KD~\cite{hinton2015distilling} has the lowest distillation time but only leads to a moderate gain. In contrast, HSAKD~\cite{yang2021hierarchical} achieves the best distillation performance, even matching the teacher accuracy, but the time is 3$\times$ than the vanilla KD. DKD~\cite{zhao2022decoupled}, a modified version of the conventional KD~\cite{hinton2015distilling}, has a desirable balance between accuracy and distillation time. From the perspective of model compression, the best-distilled ResNet-20 has 3$\times$ fewer parameters and FLOPs but only results in a 0.1\% performance drop compared to the teacher ResNet-56. In practice, we can choose a suitable KD algorithm according to your actual requirements and computing resources. 

\subsection{Online Knowledge Distillation}

\begin{figure}[tbp]  
	\centering 
	\includegraphics[width=0.8\linewidth]{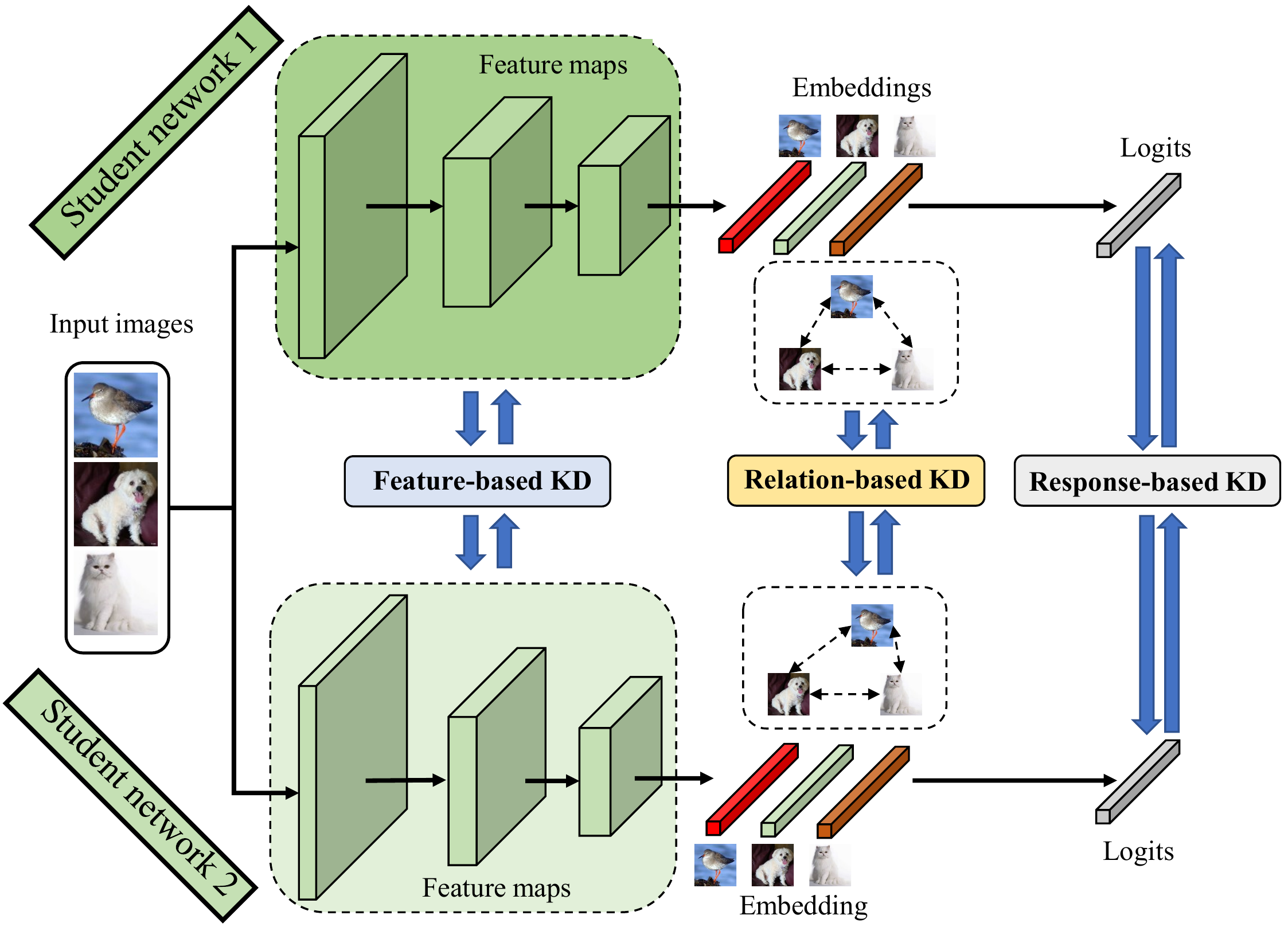}
	\caption{\textbf{The schematic illustration of online KD.} Compared with offline KD in~\ref{overview_kd}}, online KD conducts bidirectional knowledge transfer among two student networks.
	\label{online_kd_pic}
\end{figure}

Online KD aims to train a group of student networks simultaneously from scratch and transfer knowledge from each other during the training phase. Unlike offline KD, online KD is an end-to-end optimization process and does not need an explicit pre-trained teacher network in advance. According to the knowledge type, the current online KD is mainly divided into response-based, feature-based, and relation-based approaches, as illustrated in Fig.~\ref{online_kd_pic}. We provide an empirical study of various online KD approaches in Table~\ref{online_kd}.

\textbf{Response-based Online KD.} The online KD dates back to Deep mutual learning (DML)~\cite{zhang2018deep}. DML reveals that aligning each student's class posterior with that of other students learns better than training alone in the traditional learning scheme. This idea is further extended to a hierarchical architecture with shared low-level layers and separated high-level branches by Song~\emph{et al.}~\cite{song2018collaborative}. Anil~\emph{et al.}~\cite{anil2018large} applied mutual distillation to a large-scale distributed neural network. DCML~\cite{yao2020knowledge} augments mutual learning with well-designed auxiliary classifiers added into hidden layers. MutualNet~\cite{yang2020mutualnet} performs mutual learning on sub-networks equipped with different widths using various input resolutions to explore multi-scale features. MMT~\cite{ge2020mutual} and PCL~\cite{wu2021peer} introduce a temporal mean teacher for each peer to generate better pseudo-labels for mutual learning. Beyond the peer-teaching manner, ONE~\cite{zhu2018knowledge} assembles class probabilities to construct a virtual teacher role for providing soft labels. OKDDip~\cite{chen2020online} utilizes self-attention mechanism~\cite{vaswani2017attention} to boost peer diversity and then transfers the ensemble knowledge of auxiliary peers to the group leader. KDCL~\cite{guo2020online} investigates generating soft ensemble targets using various aggregation strategies from two data-augmentation views. Several works~\cite{kim2021feature,wu2021peer} consider using feature fusion with an extra classifier to output meaningful labels.

\textbf{Feature-based Online KD.} Previous online KD approaches often focus on learning class probabilities and mainly differ in various strategies or architectures, but neglect feature-level information for online learning. Walawalkar~\emph{et al.}~\cite{walawalkar2020online} performed online mimicry of intermediate feature maps for model compression. Zhang~\emph{et al.}~\cite{zhang2018deep} showed that aligning feature maps directly may diminish group diversity and harm online KD. Many works~\cite{chung2020feature,zhang2020amln} proposed online adversarial feature distillation to mutually learn feature distributions. The idea of adversarial online KD is to add a discriminator for each network that can classify the feature map from its own as fake or the other network as real. 

\textbf{Relation-based Online KD.} MCL~\cite{yang2022mutual} regards each network as an individual view and introduces mutual relation-based distillation from the perspective of contrastive representation learning. Compared with previous works, MCL~\cite{yang2022mutual} helps each network to learn better visual feature representations. A common characteristic of previous online KD methods is that distilled knowledge types are extracted from a single original task. HSSAKD~\cite{yang2022knowledge} attachs classifiers after feature maps to learn an extra self-supervision augmented task and guides networks to distill self-supervised distributions mutually.

\begin{table}[tbp]
	\centering
	\caption{Top-1 accuracy(\%) of various online KD methods for CIFAR-100~\cite{krizhevsky2009learning} classification. The compared works are sorted according to the published time. The network is selected as ResNet-32~\cite{he2016deep}. All results are referred to the original papers.}
	\resizebox{1.\linewidth}{!}{
	\begin{tabular}{l|c|c|cc}  
		\toprule
		
		Method&Venue&Algorithm&Student(baseline)&After KD\\
		\midrule
		DML~\cite{zhang2018deep}&CVPR-2018&mutual learning&ResNet-32(71.28)&73.68\\
		CL~\cite{song2018collaborative}&NeurIPS-2018&hierarchical sharing&ResNet-32(71.28)&72.33\\
		ONE~\cite{zhu2018knowledge}&NeurIPS-2018&naive ensemble learning&ResNet-32(71.28)&73.79\\
		OKDDip~\cite{chen2020online}&AAAI-2020&ensemble learning with self-attention&ResNet-32(71.28)&73.25\\
		KDCL~\cite{guo2020online}&CVPR-2020&ensemble learning with augmentation&ResNet-32(71.28)&73.76\\
		AFD~\cite{chung2020feature}&ICML-2020&adversarial feature distillation&ResNet-32(69.38)&74.03\\
		AMLN~\cite{zhang2020amln}&ECCV-2020&adversarial feature distillation&ResNet-32(69.71)&74.69\\
		PCL~\cite{wu2021peer}&AAAI-2021&mean teacher&ResNet-32(71.28)&74.14\\
		FFL~\cite{kim2021feature}&ICPR-2021&feature fusion&ResNet-32(71.28)&72.18\\
		MCL~\cite{yang2022mutual}&AAAI-2022&mutual contrastive learning&ResNet-32(70.91)&74.04\\
		HSSAKD~\cite{yang2022knowledge}&TNNLS-2022&self-supervision augmentation&ResNet-32(70.91)&74.17\\
		\bottomrule
	\end{tabular}}
	
	\label{online_kd} 
\end{table}

\subsection{Self-Knowledge Distillation}
Self-KD aims to distill knowledge explored from the network to teach itself. Unlike offline and online KD, Self-KD does not have additional teachers or peer networks for knowledge communication. Therefore, existing Self-KD works often utilize \emph{auxiliary architecture}~\cite{sun2019deeply,zhang2019your,zhang2020auxiliary,ji2021refine,zhang2021self} , \emph{data augmentation}~\cite{xu2019data,yun2020regularizing,yang2022mixskd} or \emph{sequential snapshot distillation}~\cite{furlanello2018born,yang2019snapshot,kim2021self,shen2022self} to explore external knowledge for self-boosting. Moreover, by manually designing regularization distributions to replace the teacher~\cite{yuan2020revisiting}, Self-KD can also be  connected to label smoothing~\cite{szegedy2016rethinking}. We provide an empirical study of various Self-KD approaches in Table~\ref{self_kd}. Beyond applying Self-KD over conventional supervised learning, recent works also attempt to borrow the idea of Self-KD for self-supervised learning.

\textbf{Self-KD with auxiliary architecture.} The idea of this approach is to attach auxiliary architectures to capture extra knowledge to complement the primary network. DKS~\cite{sun2019deeply} inserts several auxiliary branches and performs pairwise knowledge transfer among these branches and the primary backbone. BYOT~\cite{zhang2019your} transfers probability and feature information from the deeper portion of the network to shallow ones.  SAD~\cite{hou2019learning} uses attention maps from the deeper layer to supervise the shallow layer's ones in a layer-wise manner. Besides peer-to-peer transfer, MetaDistiller~\cite{liu2020metadistiller} constructs a label generator by fusing feature maps in a top-down manner and optimizes it with meta-learning. FRSKD~\cite{ji2021refine} aggregates feature maps in a BiFPN-like way to build a self-teacher network for providing refined feature maps and soft labels. A issue is that the auxiliary-architecture-based method highly depends on the human-designed network, and its expansibility is poor.

\textbf{Self-KD with data augmentation.} The data-augmentation-based methods often force similar predictions generated from two different augmented views. Along this vein, DDGSD~\cite{xu2019data} applies two different augmentation operators over the same image. CS-KD~\cite{yun2020regularizing} randomly samples two different instances from the same category. MixSKD~\cite{yang2022mixskd} regards the Mixup image as a view and the linearly interpolated image as the other view in the feature and probability space. To excavate cross-image knowledge, BAKE~\cite{ge2021self} attempts to absorb the other samples' knowledge via weighted aggregation to form a soft target. In general, data-augmentation-based Self-KD needs multiple forward processes compared with the baseline and improves the training costs.
\begin{table}[tbp]
	\centering
	\caption{Top-1 accuracy(\%) of various Self-KD methods for CIFAR-100~\cite{krizhevsky2009learning} classification. The compared works are sorted according to the published time. The networks are selected as ResNets~\cite{he2016deep}. All results are referred to the original papers.}
	\resizebox{1.\linewidth}{!}{
		\begin{tabular}{l|c|ccc}  
			\toprule
			Method&Venue&Algorithm&Student(baseline)&After KD\\
			\midrule
			BAN~\cite{furlanello2018born}&ICML-2018&born again&ResNet-32(68.39)&69.84\\
			DDGSD~\cite{xu2019data}&AAAI-2019&data-distortion invariance&ResNet-18(76.24)&76.61\\
			DKS~\cite{sun2019deeply}&CVPR-2019&pairwise knowledge transfer&ResNet-18(76.24)&78.64\\
			SD~\cite{yang2019snapshot}&CVPR-2019&previous snapshot&ResNet-32(68.99)&71.78\\
			BYOT~\cite{zhang2019your}&ICCV-2019&deep-to-shallow classifier&ResNet-18(76.24)&77.88\\
			SAD~\cite{hou2019learning}&ICCV-2019&layer-wise attention &ResNet-18(76.24)&76.40\\
			CS-KD~\cite{yun2020regularizing}&CVPR-2020&class-wise regularization&ResNet-18(76.24)&78.01\\
			Tf-KD~\cite{yuan2020revisiting}&CVPR-2020&manual distribution&ResNet-18(76.24)&76.61\\
			MetaDistiller~\cite{liu2020metadistiller}&ECCV-2020&meta-learning&ResNet-18(77.31)&79.05\\
			BAKE~\cite{ge2021self}&ArXiv-2021&knowledge ensembling&ResNet-18(76.24)&78.72\\
			PS-KD~\cite{kim2021self}&ICCV-2021&progressive refinement&ResNet-18(75.82)&79.18\\
			FRSKD~\cite{ji2021refine}&CVPR-2021&feature refinement&ResNet-18(76.24)&77.71\\
			DLB~\cite{shen2022self}&CVPR-2022&last mini-batch regularization&ResNet-18(73.63)&76.12\\
			MixSKD~\cite{yang2022mixskd}&ECCV-2022&Mixup regularization&ResNet-18(76.24)&80.32\\
			\bottomrule
	\end{tabular}}
	\label{self_kd} 
\end{table}

\textbf{Self-KD with sequential snapshot distillation.} This vein considers making use of the network's counterparts along the training trajectory to provide supervisory signals. BAN~\cite{furlanello2018born} gradually improves the network under the supervision of previously trained counterparts in a sequential manner. SD~\cite{yang2019snapshot} takes the network snapshots from earlier epochs to teach its later epochs. PS-KD~\cite{kim2021self} proposes progressively refining soft targets by summarizing the ground-truth and past predictions. DLB~\cite{shen2022self} performs consistency regularization between the last and current mini-batch. In general, snapshot-based Self-KD requires saving multiple copies of the training model and increases memory costs.

\textbf{Self-supervised learning with Self-KD.} Self-supervised learning focuses on good feature representations given unannotated data. There are some interesting connections between self-KD and self-supervised learning. In the self-supervised scenario, the framework often constructs two roles: \emph{online} and \emph{target} networks. The former is the training network, and the latter is a mean teacher~\cite{tarvainen2017mean} with moving-averaged weights from the online network. The target network has an identical architecture
to the online network but has different weights. The target network is often used to provide supervisory signals to train the online network. MoCo~\cite{he2020momentum} utilizes the target network to generate consistent positive and negative contrastive samples. Some self-supervised works regard the target network as a self-teacher to provide the regression targets, such as BYOL~\cite{grill2020bootstrap}, DINO~\cite{caron2021emerging} and SimSiam~\cite{chen2021exploring}. Inspired by BYOT~\cite{zhang2019your}, SDSSL~\cite{jang2021self} guides the intermediate feature embeddings to contrast the features from the final layer. Although previous methods achieve desirable performance on self-supervised representation learning, there may still exist two directions to be worth exploring. First, the target network is constructed from the online network in a moving-averaged manner. Do we have a more meaningful way to build the target network? Second, the loss is often performed to align the final feature embeddings. Some intermediate features or contrastive relationships between the online and target networks could be further mined.

\begin{table}[tbp]
	\centering
	\caption{Comprehensive comparison of three knowledge distillation schemes. '*' denotes that offline KD has publicly available pre-trained teacher models at hand.}
	\begin{tabular}{l|c|c|c|c}  
		\toprule
		Aspect&Offline KD&Offline KD (*)&Online KD&Self-KD\\
		\midrule
		end-to-end training&no&yes&yes&yes\\
		extra models&yes&yes&yes&no\\
		computational complexity&large&low&medium&low\\
		\bottomrule
	\end{tabular}
	\label{Comprehensive} 
\end{table}
\subsection{Comprehensive Comparison}
As shown in Table~\ref{Comprehensive}, we comprehensively compare three KD schemes in various aspects. Offline KD needs an extra teacher model to train a student, while online KD or Self-KD trains a cohort of models or a single model with an end-to-end optimization. When publicly available pre-trained teacher models are unavailable, pre-training a teacher network for offline KD often has high capacity and thus is time-consuming for training. Self-KD utilizes a single model for self-boosting and often has low computational complexity. It is noteworthy that offline KD with publicly available pre-trained teacher models also has low complexity because the inference time for extra frozen teacher models without gradient-propagation does not introduce much cost.

\section{Distillation Algorithms}
\subsection{Multi-Teacher Distillation}

\begin{figure}[tbp]  
	\centering 
	\includegraphics[width=0.8\linewidth]{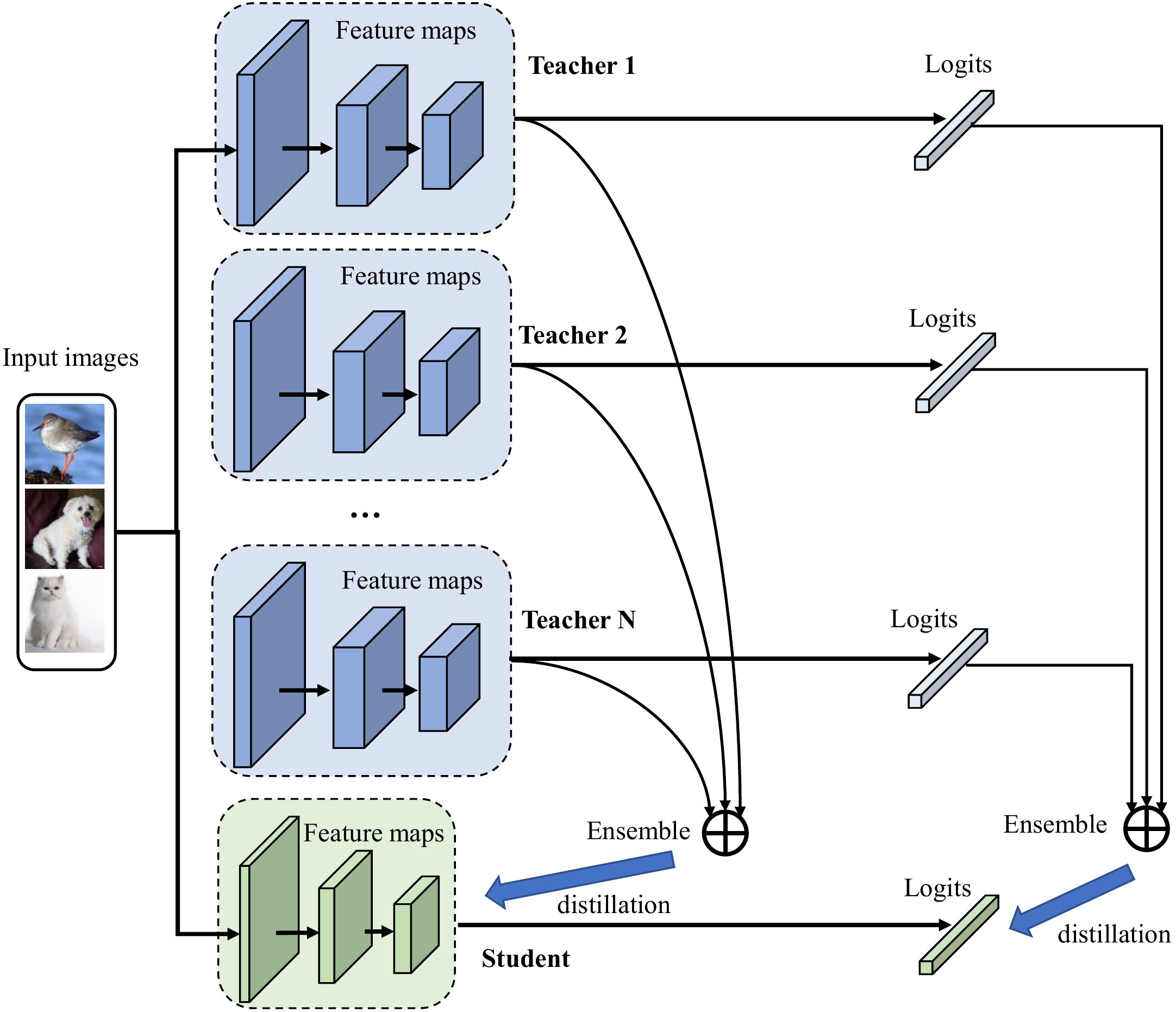}
	\caption{\textbf{The schematic illustration of multi-teacher distillation.} This framework assembles multiple teachers' feature maps and logits to distill a single student network.} 
	\label{multi_kd}
\end{figure}
In the traditional KD, knowledge is transferred from a high-capacity teacher to a compact student. But in this setting, the knowledge diversity and capacity are limited. Different teachers can provide their unique and useful knowledge to student. In this way, the student can learn various knowledge representations from multiple teacher networks. Following the vanilla KD, the knowledge in the form of logits or the intermediate features can be used as a supervision signal. The schematic illustration of multi-teacher KD is shown in Fig.~\ref{multi_kd}.

\textbf{KD from ensemble logits}. Logits from the model ensemble are one of the direct ways in multi-teacher knowledge distillation. Based on this idea, the student is guided to learn the soft output of teachers' ensemble logits in ~\cite{mirzadeh2020improved,you2017learning}. However, the simple average of individual predictions may ignore the importance of variety and diversity among the teacher group. Therefore, some works ~\cite{fukuda2017efficient,xiang2020learning} proposed learning the student model by adaptively imitating teachers' outputs with various aggregation weights. 

\textbf{KD from ensemble feature representations}. Besides distilling from logits, the ensemble of intermediate representations ~\cite{park2019feed,wu2019distilled,liu2019knowledge,chen2019two} can provide more semantic information to student. However, it is more challenging for distillation from feature representations, since each member in ensemble teachers has various feature representations in specific layers. To address this issue, Park~\emph{et al.}~\cite{park2019feed} applied non-linear transformations to multiple teacher networks at the feature-map level. Wu~\emph{et al.}~\cite{wu2019distilled} proposed to distill the knowledge by minimizing the distance between the similarity matrices of teachers and a student. Liu~\emph{et al.}~\cite{liu2019knowledge} proposed to let the student network learn the teacher models' learnable transformation matrices. To take advantage of both logits and intermediate features, Chen~\emph{et al.}~\cite{chen2019two} introduced double teacher networks to provide response-level and feature-level knowledge, respectively.

\textbf{Computation-efficient multi-teacher KD from sub-networks}. Using multi-teacher introduces extra training computation costs and decreases the training process. Thus some methods ~\cite{you2017learning,nguyen2021stochasticity} create some sub-teachers from a single teacher network. Nguyen~\emph{et al.}~\cite{nguyen2021stochasticity} utilized stochastic blocks and skipped connections over a teacher network to produce several teacher roles. Several works~\cite{song2018collaborative,he2018multi} designed multi-headed architectures to produce many teacher roles.

\textbf{Multi-task multi-teacher KD}. In most cases, multi-teacher KD is based on the same task. Knowledge amalgamation~\cite{shen2019customizing} is proposed to learn a versatile student via learning knowledge from all teachers trained by different tasks. Luo \emph{et al.}~\cite{luo2019knowledge} aimed to learn a multi-talented student network that can absorb comprehensive knowledge from heterogeneous teachers. Ye \emph{et al.}~\cite{ye2019amalgamating} focused a target network for customized tasks guided by multiple teachers pre-trained from different tasks. The student inherits desirable capabilities from heterogeneous teachers so that it can perform multiple tasks simultaneously. Rusu~\emph{et al.}~\cite{rusu2015policy} introduced a multi-teacher policy distillation approach to transfer agents' multiple policies to a single student network.

\textbf{Discussion.} In summary, a versatile student can be trained via multi-teacher distillation since different teachers provide diverse knowledge. However, several problems still deserve to be solved. On the one hand, the number of teachers is a trade-off problem between training costs and performance improvements. On the other hand, integrating various knowledge from multiple teachers effectively is still an open issue.

\subsection{Cross-Modal Distillation}
\begin{figure}[tbp]  
	\centering 
	\includegraphics[width=0.6\linewidth]{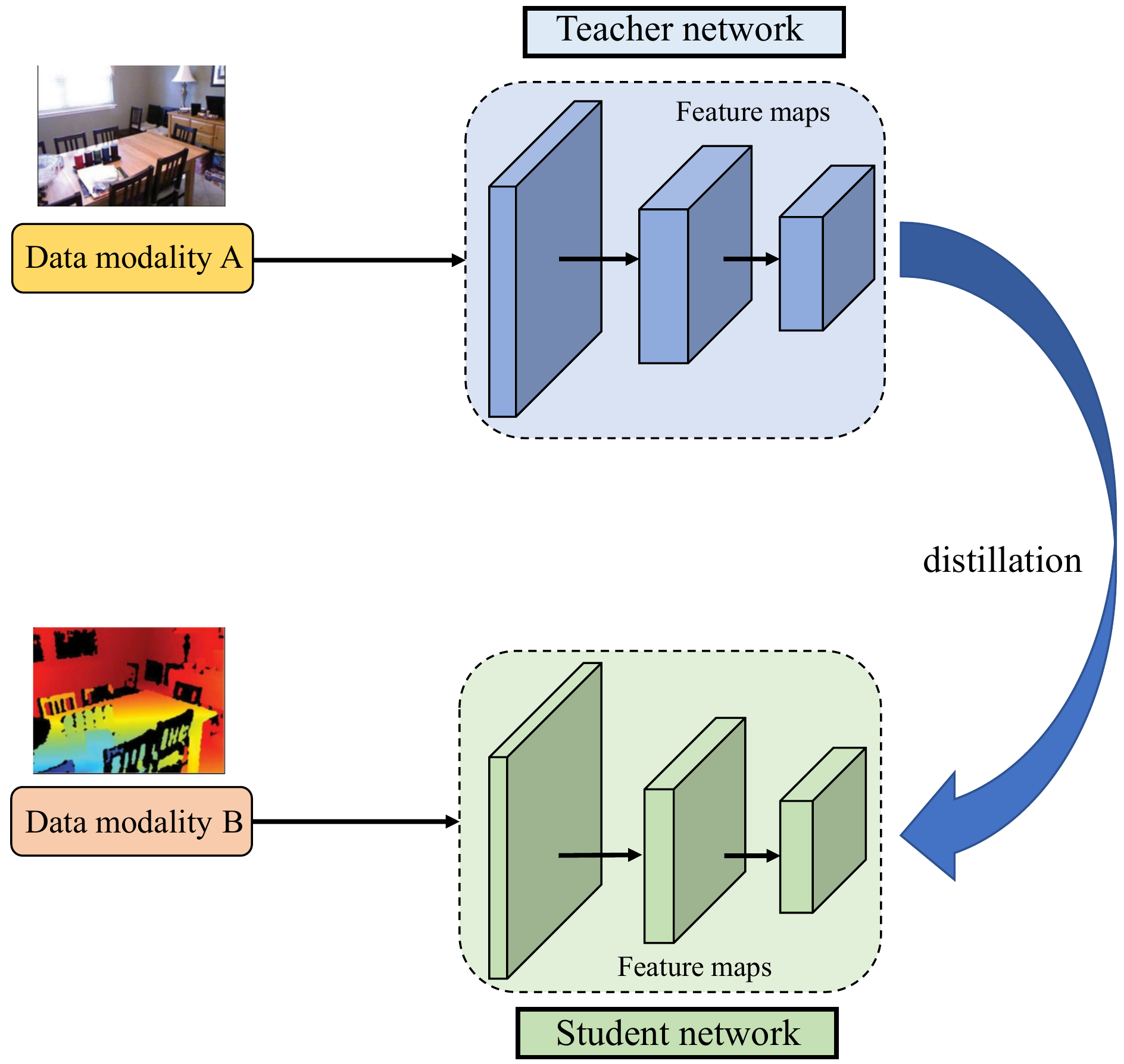}
	\caption{\textbf{The schematic illustration of cross-modal distillation.} It transfers knowledge from the teacher trained by a data modality to a student network from another data modality. } 
	\label{cross_modal_kd}
\end{figure}

The teacher and student in common KD methods often have the same modality. However, the training data or labels for another modalities may be unavailable. Transferring knowledge between different modalities is a valuable field in practice. The core idea of cross-modal KD  is to transfer knowledge from the teacher trained by a data modality to a student network from another data modality. The schematic illustration of cross-modal KD is shown in Fig.~\ref{cross_modal_kd}.

 Given a teacher model pre-trained on one modality with well-labeled samples, Gupta~\emph{et al.}~\cite{gupta2016cross} transfered information between annotated RGB images and unannotated optical flow images leveraging unsupervised paired samples. The paradigm via label-guided pair-wise samples has been widely applied for cross-modal KD. Thoker~\emph{et al.}~\cite{thoker2019cross} transferred knowledge from RGB videos to a 3D human action recognition model using paired samples. Roheda~\emph{et al.}~\cite{roheda2018cross} proposed cross-modality distillation from an available modality to a missing modality using GANs. Do~\emph{et al.}~\cite{do2019compact} explored a KD-based visual question answering method and relied on supervised learning for cross-modal transfer using the ground truth labels. Passalis~\emph{et al.}~\cite{passalis2018learning} proposed probabilistic KD to transfer knowledge from textual modality into visual modality.

\textbf{Discussion}. Generally, KD performs well on cross-modal scenarios. However, cross-modal KD is difficult to model knowledge interaction when a significant modality gap exists.

\subsection{Attention-based Distillation}
\begin{figure}[tbp] 
	\centering 
	\subfigure[Attention maps]{ 
		\label{att_maps} 
		\includegraphics[width=0.57\textwidth]{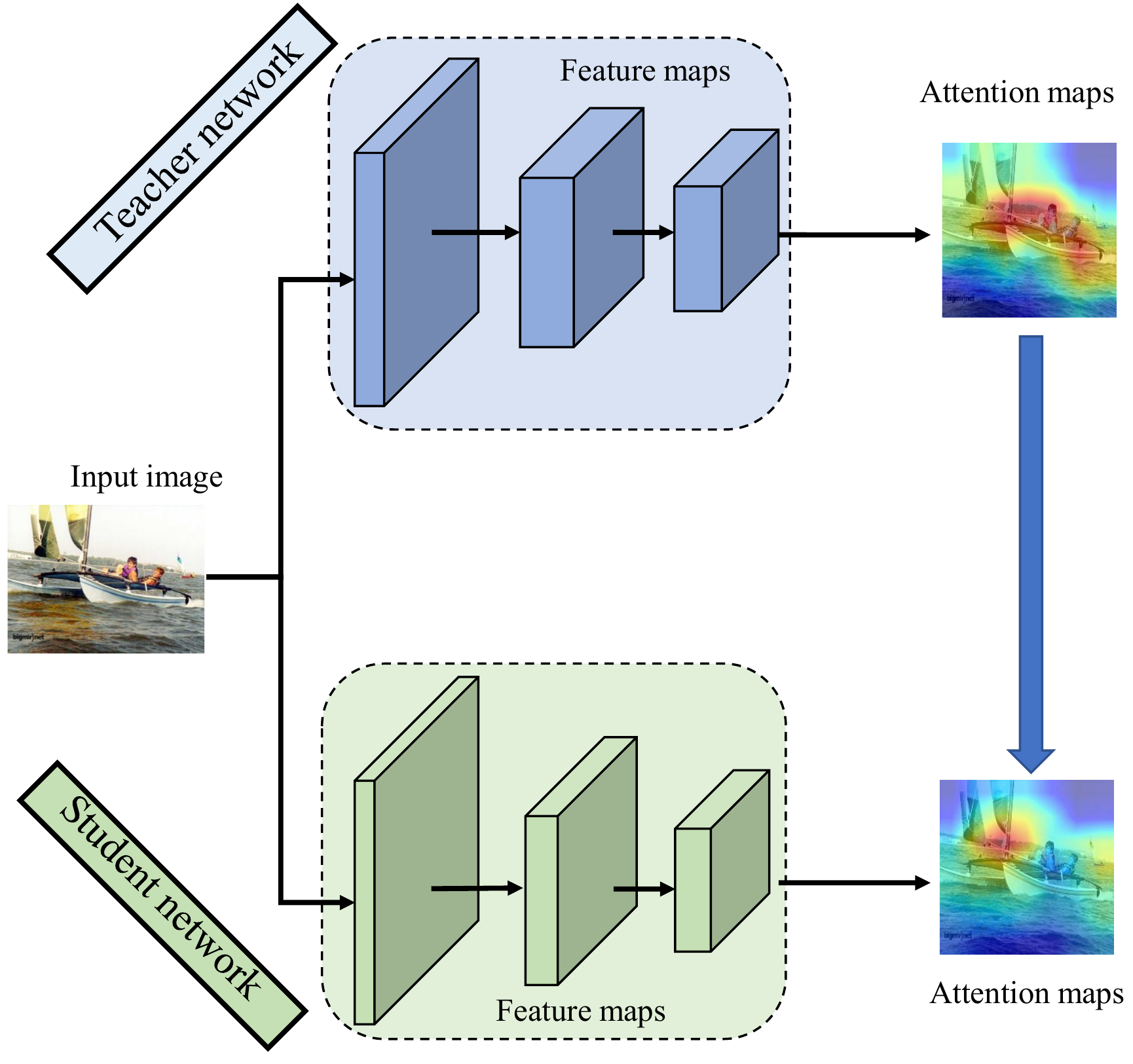}} 
	\subfigure[Attention weights]{ 
		\label{att_weights} 
		\includegraphics[width=0.4\textwidth]{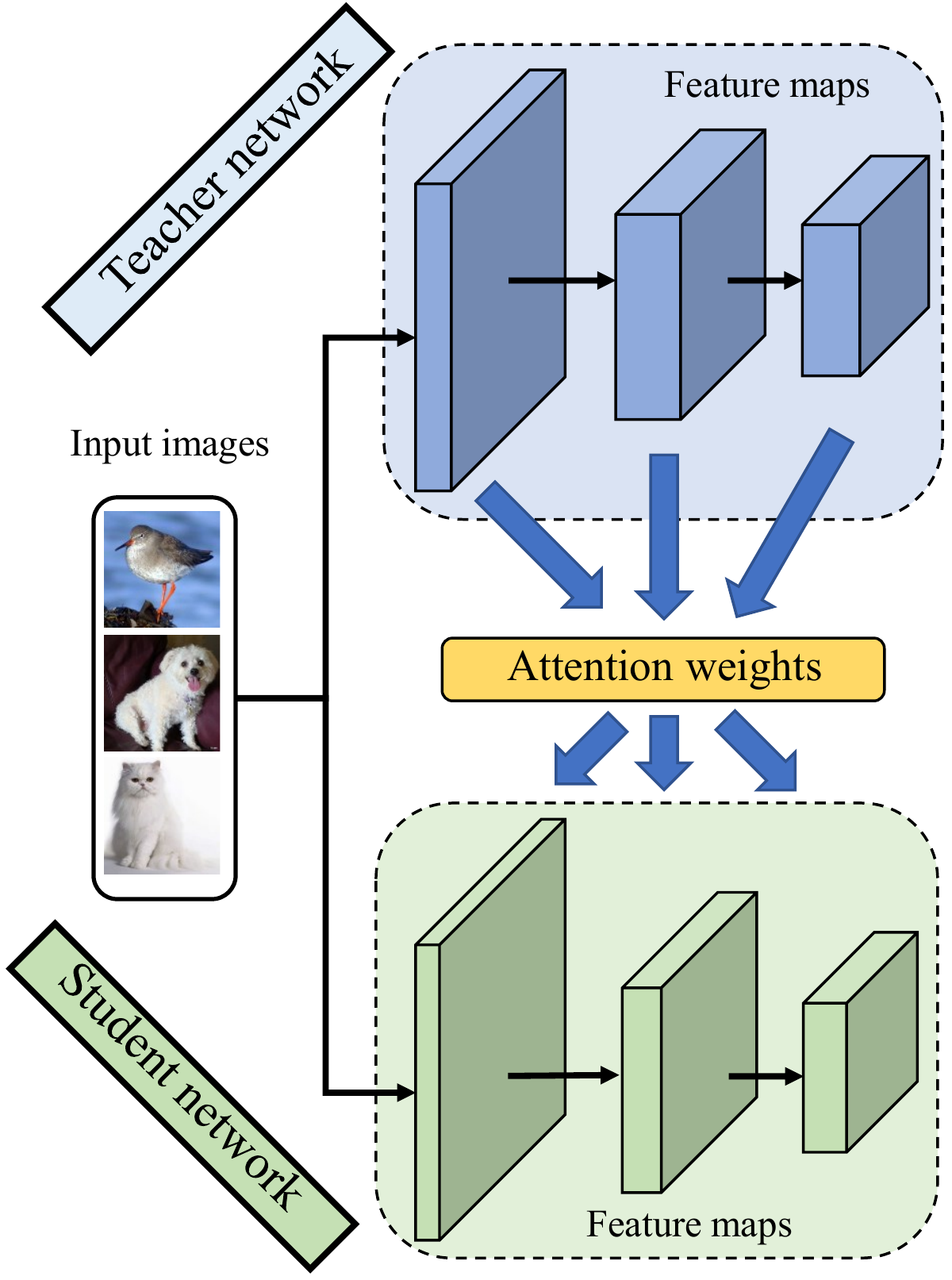}} 
	\caption{\textbf{Attention-based distillation with (a) maps or (b) weights.} } 
	\label{attention_kd} 
\end{figure}

Attention-based distillation takes advantage of attention information for effective knowledge transfer. Current works follow two veins: (1) distilling attention maps refined from feature maps, and (2) weighted distillation based on self-attention mechanism~\cite{vaswani2017attention}, as illustrated in Fig.~\ref{attention_kd}.

\textbf{Distilling attention maps}. Attention maps often reflect valuable semantic information and suppress unimportant parts. The seminal AT~\cite{zagoruyko2016paying} constructs a spatial attention map by calculating statistics from a feature map across the channel dimension and performs alignment of attention maps between the teacher and student networks. The spatial attention map contains class-aware semantic regions that help the student to capture discriminative features. CD~\cite{zhou2020channel} employs a squeeze-and-excitation module~\cite{hu2018squeeze} to generate channel attention maps and lets the student learn the teacher's channel attention weights. CWD~\cite{shu2021channel} distills a spatial attention map per channel representing semantic masks for dense predictions. TinyBert~\cite{jiao2019tinybert} transfers self-attention matrices for transformer-layer distillation. LKD~\cite{li2020local} introduces a class-aware attention module to capture class-relevant regions for constructing a local correlation matrix.

\textbf{Self-attention-based weighted distillation}. The self-attention technique is a desirable mechanism to capture similarity relationships~\cite{vaswani2017attention} among features. Several works~\cite{chen2021cross,ji2021show,passban2021alp}
applied attention-based weights for adaptive layer-to-layer semantic matching. SemCKD~\cite{chen2021cross} automatically assigns targets aggregated from suitable teacher layers with attention-based similarities for each student layer. AFD~\cite{ji2021show} proposes an attention-based meta-network to model relative similarities between teacher and student features. ALP-KD~\cite{passban2021alp} fuses teacher-side information with attention-based layer projection for Bert~\cite{devlin2018bert} distillation. Orthogonal to the layer-wise assignment, TTKD~\cite{lin2022knowledge} applies self-attention mechanism~\cite{vaswani2017attention} for spatial-level feature matching.

\textbf{Discussion.} The attention-based map captures saliency regions and filters redundant information, helping the student learn the most critical features. However, the condensed attention map compresses the feature map's dimension and may lose meaningful knowledge. Moreover, the attention map may sometimes not focus on the correct region, resulting in an adverse supervisory impact. 
\subsection{Data-free Distillation}

\begin{figure}[tbp]  
	\centering 
	\includegraphics[width=0.8\linewidth]{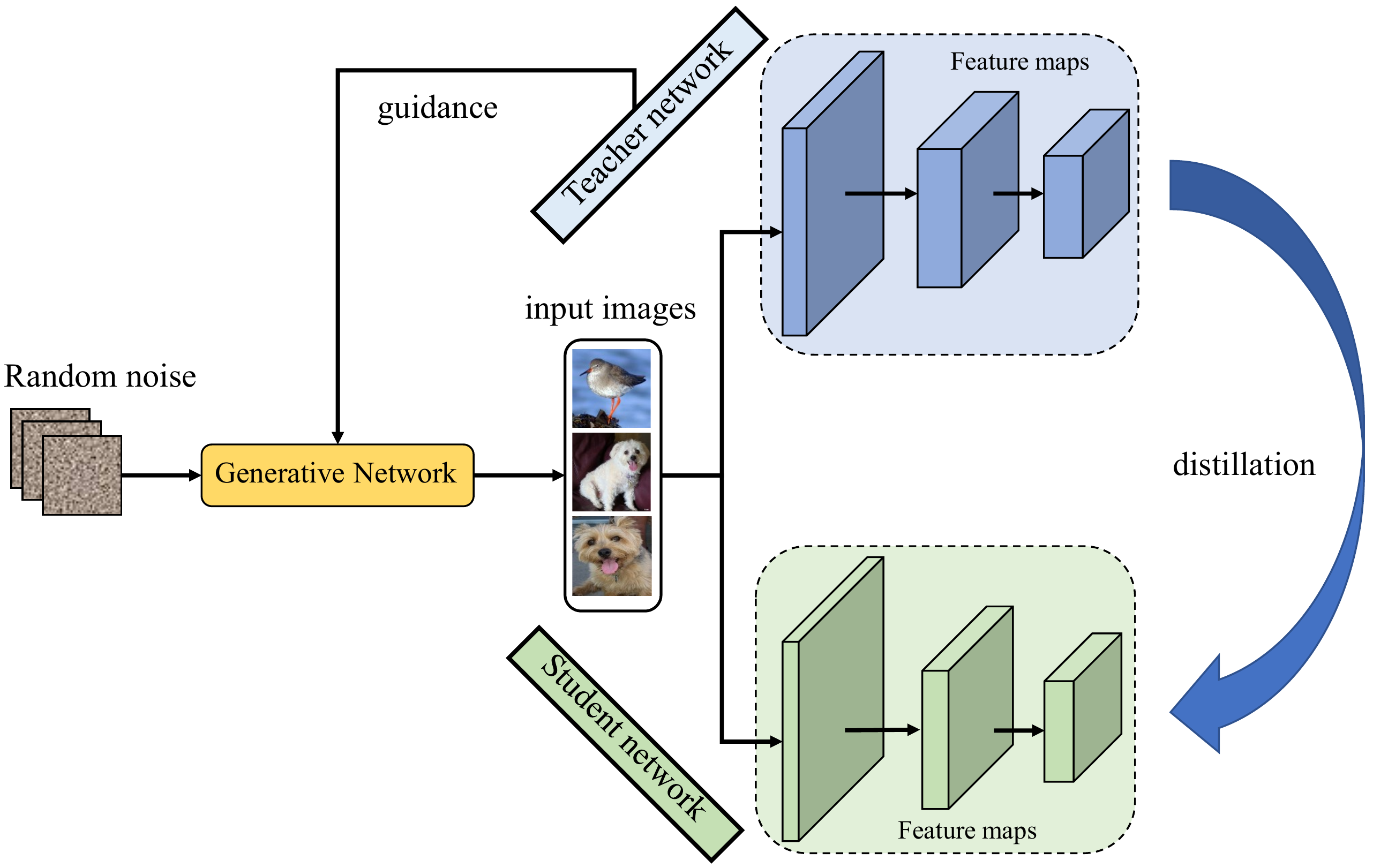}
	\caption{\textbf{The schematic illustration of data-free distillation.} It relies on a generative network to produce images from random noise as input for distillation.} 
	\label{data_free_kd}
\end{figure}

The traditional KD methods often need large training samples. However, the training dataset may sometimes be unavailable due to privacy or safety concerns. Some methods have been proposed to handle this problem, mainly divided into data-free KD and dataset KD. The schematic illustration of data-free KD is shown in Fig.~\ref{data_free_kd}.

\textbf{Data-free KD.} Training samples are often newly or synthetically produced using generative adversarial networks (GAN)~\cite{goodfellow2014generative}. The teacher network supervises the student network with generated samples as input. Lopes~\emph{et al.}~\cite{lopes2017data} used different types of activation records to reconstruct the original data. DeepInversion~\cite{yin2020dreaming} explores information stored in batch normalization layers to generate synthesized samples for data-free KD. Nayak~\emph{et al.}~\cite{nayak2019zero} introduced a sample extraction mechanism by modeling the softmax space as a Dirichlet distribution from the teacher's parameters. Beyond the final output, the target data can be generated using the information from the teacher's feature representations~\cite{micaelli2019zero,nayak2019zero}. Paul~\emph{et al.}~\cite{micaelli2019zero} optimized an adversarial generator to search for difficult images and then used these images to train the student. CMI~\cite{fang2021contrastive} introduces contrastive learning to make the synthesizing instances distinguishable compared to already synthesized ones. FastDFKD~\cite{fang2022up} optimizes a meta-synthesizer to reuse the shared common features for faster data-free KD. Similar to zero-shot learning, some KD methods with few-shot learning were proposed to distill data-free knowledge from a teacher model into a student network~\cite{kimura2018few,shen2021progressive}, where the teacher network often uses little labeled data. 

\textbf{Dataset KD.} Besides data-free KD, dataset distillation~\cite{radosavovic2018data,liu2019ddflow} is an essential direction for synthesizing a small dataset to represent the original full dataset without much accuracy degradation. To exploit the omni-supervised setting,  Radosavovic~\emph{et al.}~\cite{radosavovic2018data} assembles predictions from multiple transformations of unlabeled data by a single model to produce new training annotations. DDFlow~\cite{liu2019ddflow} proposed to learn optical flow estimation  and distill predictions from a teacher model to supervise a student network for optical flow learning. Unlabeled data may hinder Graph Convolutional Network(GCN) from learning graph-based data. In general, pseudo labels of the unlabeled data can provide extra supervision to train GCN. RDD~\cite{zhang2020reliable} proposed a reliable data-driven semi-supervised GCN training method. It can better use high-quality data and improve graph representation learning by defining node and edge reliability. Cazenavette~\emph{et al.}~\cite{cazenavette2022dataset} conducted long-range parameter matching along training trajectories between distilled synthetic and real data.

\textbf{Discussion.}  In most data-free KD methods, the synthesis data is usually generated from the pre-trained teacher network's feature representations. Although current data-free KD works have shown remarkable performance in handling the data-unavailable issue, generating more high-quality and diverse training samples is still a challenge to be studied.

\subsection{Adversarial Distillation}
\begin{figure}[tbp] 
	\centering 
	\subfigure[Adversarial distillation]{ 
		\label{adv_kd1} 
		\includegraphics[width=0.46\textwidth]{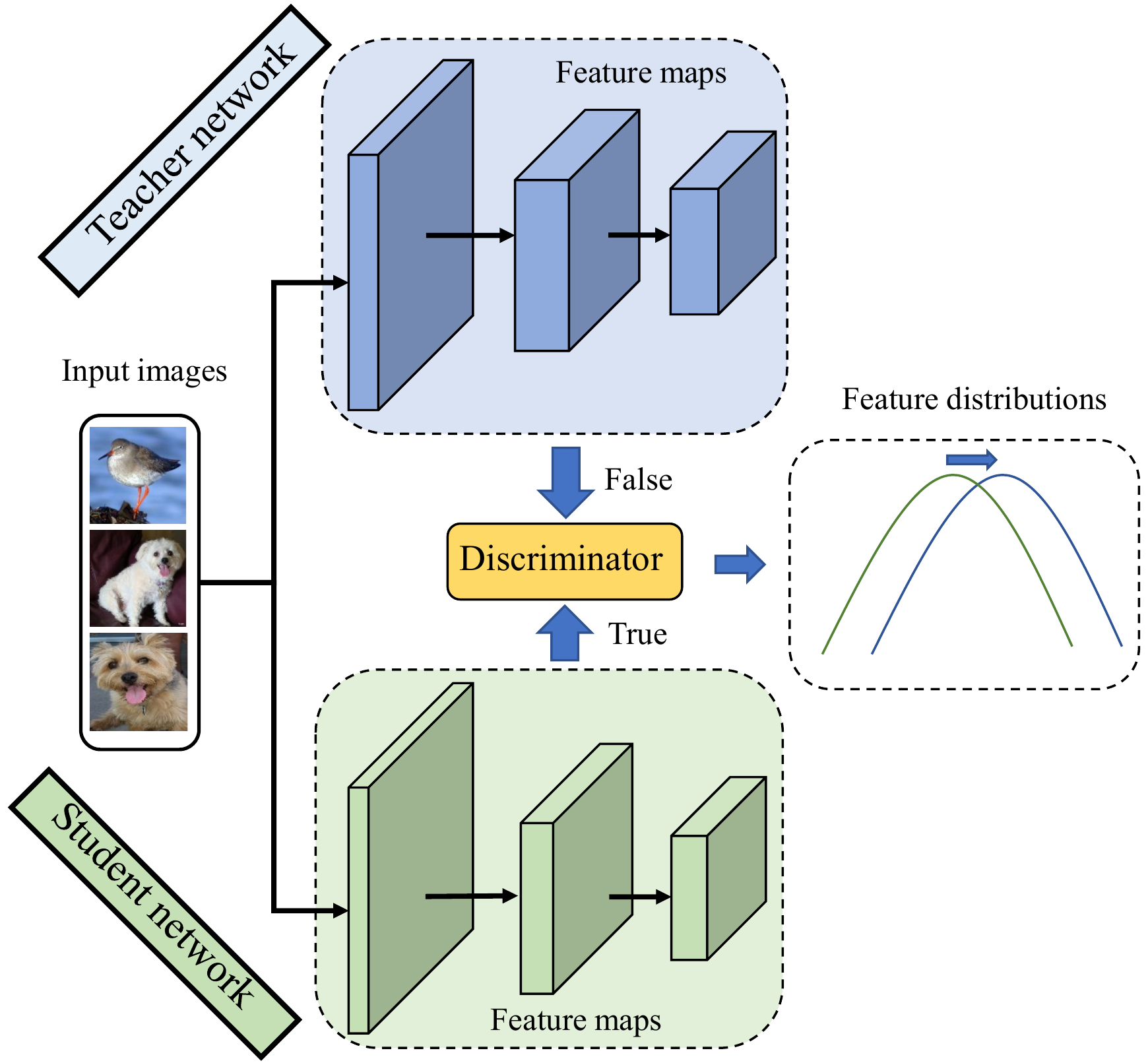}} 
	\subfigure[GAN compression]{ 
		\label{adv_kd2} 
		\includegraphics[width=0.5\textwidth]{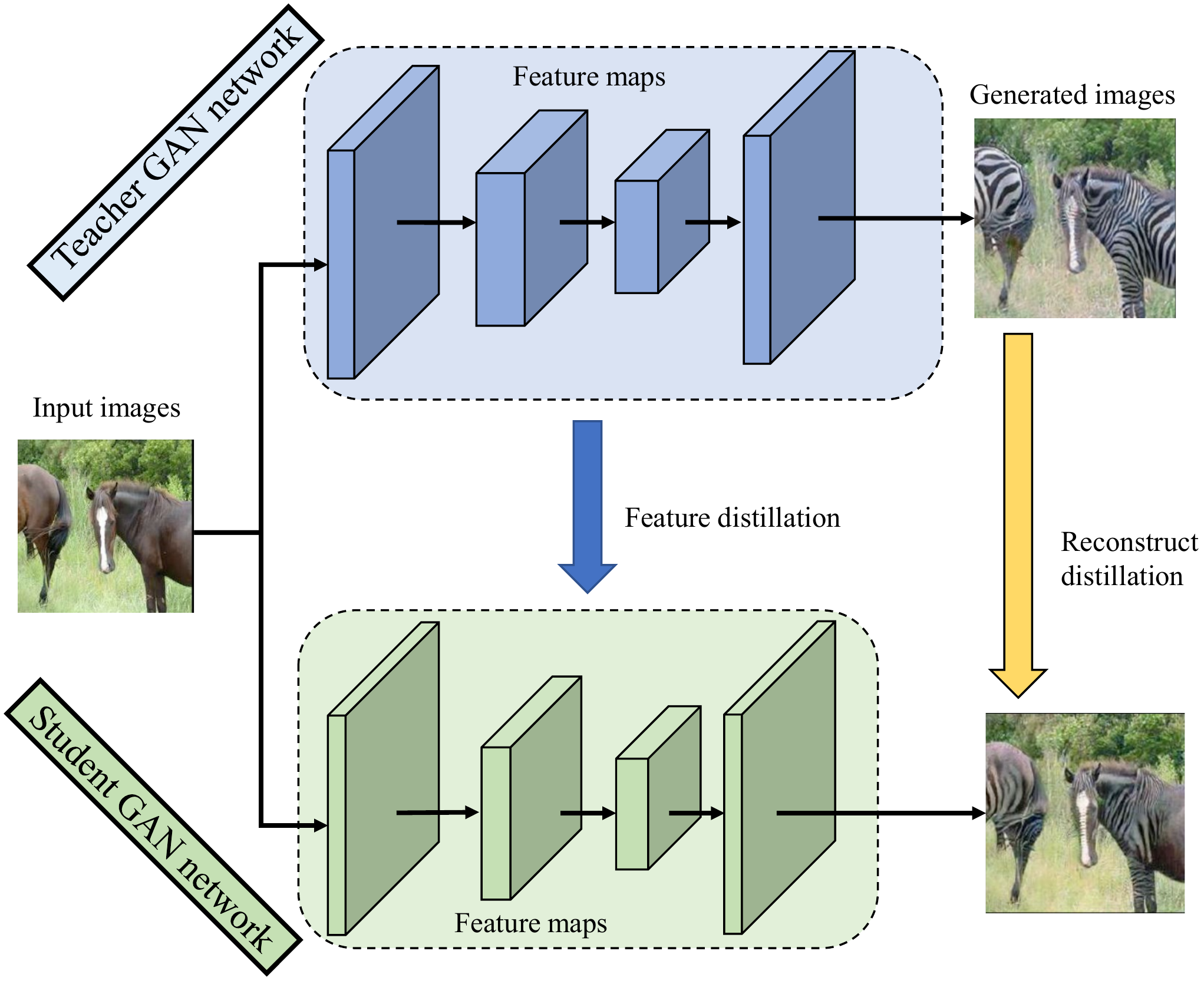}} 
	\caption{\textbf{The schematic illustrations of (1) KD with adversarial mechanism and (2) Generative Adversarial Networks (GAN) compression.}} 
	\label{adv_kd} 
\end{figure}

Adversarial distillation is to use the basic idea of generative adversarial networks to improve KD, mainly divided into three veins: (1) using GAN to generate extra data samples, (2) adversarial mechanism to assist general KD, and (3) compressing GAN for efficient image generation. The schematic illustrations of adversarial KD are shown in Fig.~\ref{adv_kd}.

\textbf{Using GAN to generate extra data samples.} Almost existing KD algorithms are data-driven, i.e., relying on original or alternative data, which may be unavailable in real-world scenarios. Generative adversarial networks can be applied to learn the true data distribution and solve this problem. DFAD~\cite{fang2019data} lets the teacher and student networks play the role of discriminator jointly to reduce the discrepancy. Meanwhile, it adds an extra generator to produce hard samples to enlarge it adversarially.
 Several works~\cite{yoo2019knowledge,liu2018teacher,roheda2018cross} introduce Conditional GAN (CGAN) to generate data. Roheda~\emph{et al.}~\cite{roheda2018cross} used CGAN to distill knowledge from missing modalities given other available modalities. Lifelong-GAN~\cite{zhai2019lifelong} transfers learned knowledge from previous networks to the new network for continual conditional image generation. 

\textbf{Adversarial mechanism to assist general KD.} The conventional KD often reduces the gap between teacher and student by aligning knowledge distributions. The adversarial mechanism can be regarded as an auxiliary method to improve the mimicry difficulty. In general, the core idea is to introduce an extra discriminator to classify feature representations from teacher or student network~\cite{belagiannis2018adversarial,wang2018adversarial,wang2020gan,liu2020ktan}. Wang~\emph{et al.}~\cite{wang2018adversarial} leveraged a discriminator as the teaching assistant to make the student learn similar feature distributions with the teacher for image classification. Wang~\emph{et al.}~\cite{wang2020gan} employed adversarial KD for one-stage object detection. Liu~\emph{et al.}~\cite{liu2019structured} transferred pixel-wise class probabilities adversarially for semantic segmentation. Beyond teacher-student-based adversarial learning, several works~\cite{chung2020feature,zhang2020amln} apply online adversarial KD for distilling feature maps mutually among multiple student networks.

 \textbf{Compressing GAN for efficient image generation.} Aguinaldo~\emph{et al.}~\cite{aguinaldo2019compressing} guided a smaller "student" GAN to align
 a larger "teacher" GAN with mean squared loss. Chen~\emph{et al.}~\cite{chen2020distilling} let a student generator learn low- and high-level knowledge from the corresponding teacher. Moreover, the student discriminator is supervised by the teacher network via triplet loss. Li~\emph{et al.}~\cite{li2020gan} proposed a general compression framework for conditional GANs by transferring knowledge from intermediate representations and exploring efficient architectures via neural architecture search. Zhang~\emph{et al.}~\cite{zhang2022wavelet} pointed out that small GANs are often difficult to generate desirable high-frequency information. WaveletKD~\cite{zhang2022wavelet} resolves images into various frequency bands via discrete wavelet transformations and then only transfers the valuable high-frequency bands.

\textbf{Discussion.} Although adversarial-based KD facilitates knowledge mimicry, it may be hard to ensure the convergence of GAN-based networks in practice. For GAN compression, what information extracted from features suitable for distilling GAN is still an open issue. 

\section{Conclusion}
In this chapter, we first survey the conventional offline KD works according to their extracted knowledge types, \emph{i.e.} response, feature and relation. Benefiting the comprehensive supervision from teacher, the student could generalize better over the target task. The teacher-student-based KD has some limitations, for example, high costs for pre-training a large teacher network. Therefore, two Online KD and Self-KD schemes are proposed to improve the student without a pre-trained teacher. In practice, KD applications often face various scenarios, for example, cross-modal KD and data-free KD. Moreover, we also show some popular mechanisms to help distillation perform better, such as multi-teacher KD, attention-based KD and adversarial KD. We survey the representative works for each KD setup and summarize their main ideas and contributions. Finally, we prospect the future challenges of existing KD applications. Compared with previously seminal KD survey papers~\cite{gou2021knowledge,wang2021knowledge}, our paper includes some newer works published at 2022 and introduces some advanced KD directions, for example, KD for vision transformer and self-supervised learning. We hope our survey can inspire future research to develop more advanced KD algorithms for improving the student performance.

\clearpage
%
%
\bibliographystyle{splncs04}
\bibliography{egbib}
\end{document}